\documentclass[10pt,letterpaper]{article}
\pdftrailerid{}
\usepackage[T1]{fontenc}
\usepackage[letterpaper,left=1.05in,right=1.05in,top=.70in,bottom=.77in,footskip=23pt]{geometry}
\usepackage{amsmath,amssymb,amsfonts,amsthm}
\usepackage{newtxtext,newtxmath}
\usepackage[scaled=.94]{helvet}
\usepackage{microtype,graphicx,booktabs,array,multirow,float,adjustbox,placeins}
\usepackage[table]{xcolor}
\usepackage{tcolorbox}
\usepackage{caption}
\usepackage[numbers,sort&compress]{natbib}
\usepackage[ruled,vlined]{algorithm2e}
\usepackage{titlesec,fancyhdr}
\usepackage[colorlinks,allcolors=linkblue,bookmarksnumbered,pdfdisplaydoctitle]{hyperref}
\definecolor{ink}{HTML}{151B2B}
\definecolor{linkblue}{HTML}{314D9A}
\definecolor{quiet}{HTML}{515969}
\definecolor{rulegray}{HTML}{CBD1DB}
\definecolor{coverink}{HTML}{1C2B34}
\definecolor{abstractfill}{HTML}{F1F4F6}
\definecolor{oursrow}{HTML}{EAF2FC}
\hypersetup{pdftitle={MOSAIC-VLA: Merged Visual Tokens Need Their Weight Back},
pdfauthor={Jinghao Zhou, Qichen Tan, Xiaojie Zhang, Ziyi Liu, Jiajie Xu, He Huang, Xing Gao, Min Zhang, Zhijun Li},
pdfsubject={Vision-language-action policies and visual token merging}}

\newcommand{\corr}{\ensuremath{{+}\!\log|G|}}
\newcommand{\runin}[1]{\par\vspace{2pt}\noindent\textbf{#1}}

\newcommand{\appendices}{\appendix}

\titleformat{\section}{\Large\sffamily\bfseries\color{ink}}{\color{linkblue}\thesection}{.65em}{}
\titleformat{\subsection}{\large\sffamily\bfseries\color{ink}}{\thesubsection}{.65em}{}
\titlespacing*{\section}{0pt}{2.1ex plus .6ex minus .3ex}{.65ex}
\titlespacing*{\subsection}{0pt}{1.65ex plus .4ex minus .25ex}{.45ex}
\newcommand{\aff}[1]{\textsuperscript{#1}}
\DeclareCaptionFont{covercaption}{\fontsize{9.55}{11.7}\selectfont}
\newcommand{\coversans}{\fontfamily{SourceSansPro-TLF}\selectfont}
\newcommand{\schoolaffiliation}[4]{%
  \mbox{\aff{#1}\,\href{#2}{%
    \raisebox{-.6pt}{#3}%
    \hspace{.28em}\textcolor{quiet}{#4}}}}
\fancypagestyle{cover}{\fancyhf{}\fancyfoot[C]{\small\color{quiet}\thepage}}
\begin{document}
\thispagestyle{cover}
\noindent\href{https://www.insightos.cn/}{\includegraphics[height=23.5pt]{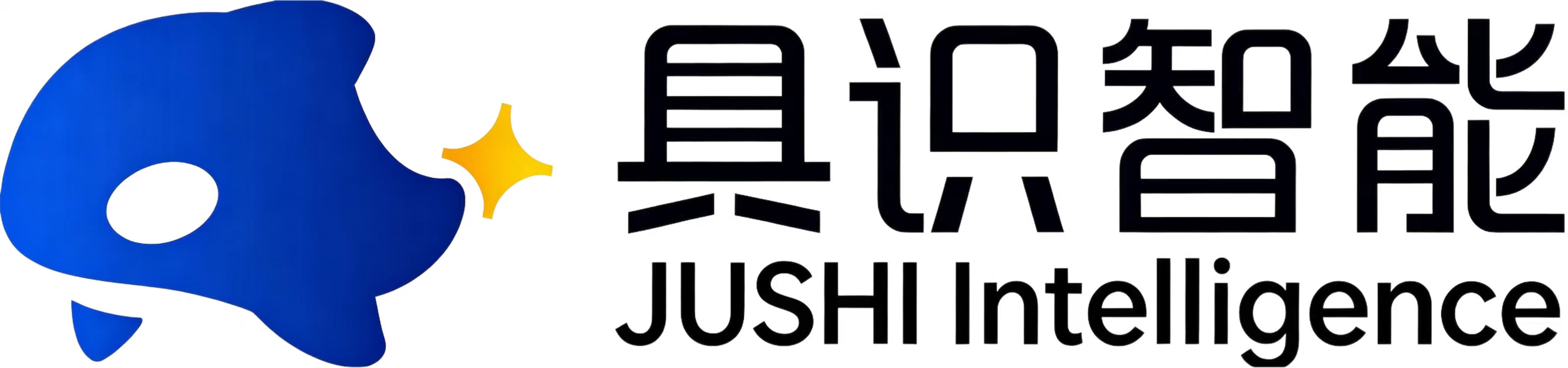}}\par
\vspace{10pt}
{\noindent\coversans\fontseries{sb}\fontsize{18.05}{21.5}\selectfont
\color{coverink}MOSAIC-VLA: Merged Visual Tokens Need Their Weight Back\par}
\vspace{6pt}{\color{rulegray}\hrule height .45pt}
\vspace{7pt}
{\noindent\coversans\fontseries{sb}\fontsize{9.6}{12}\selectfont\color{coverink}%
\href{mailto:24B903082@stu.hit.edu.cn}{\textcolor{coverink}{Jinghao Zhou}}\aff{1}, Qichen Tan\aff{2,\textdagger}, Xiaojie Zhang\aff{1}, Ziyi Liu\aff{2},
Jiajie Xu\aff{2}, He Huang\aff{2}, Xing Gao\aff{3}, Min Zhang\aff{1}, Zhijun Li\aff{1,\textdagger}\par}
\vspace{3pt}
{\noindent\fontsize{9.25}{11.5}\selectfont\color{quiet}%
\schoolaffiliation{1}{https://en.hit.edu.cn/}{\includegraphics[height=7.2pt]{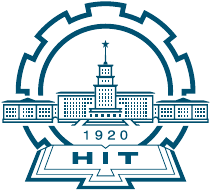}}{Harbin Institute of Technology}\qquad
\schoolaffiliation{2}{https://www.suda.edu.cn/}{\includegraphics[height=7.2pt]{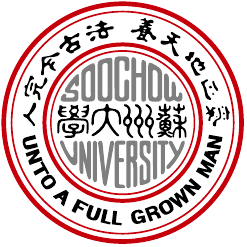}}{Soochow University}\qquad
\schoolaffiliation{3}{https://www.xmu.edu.cn/}{\includegraphics[height=7.2pt]{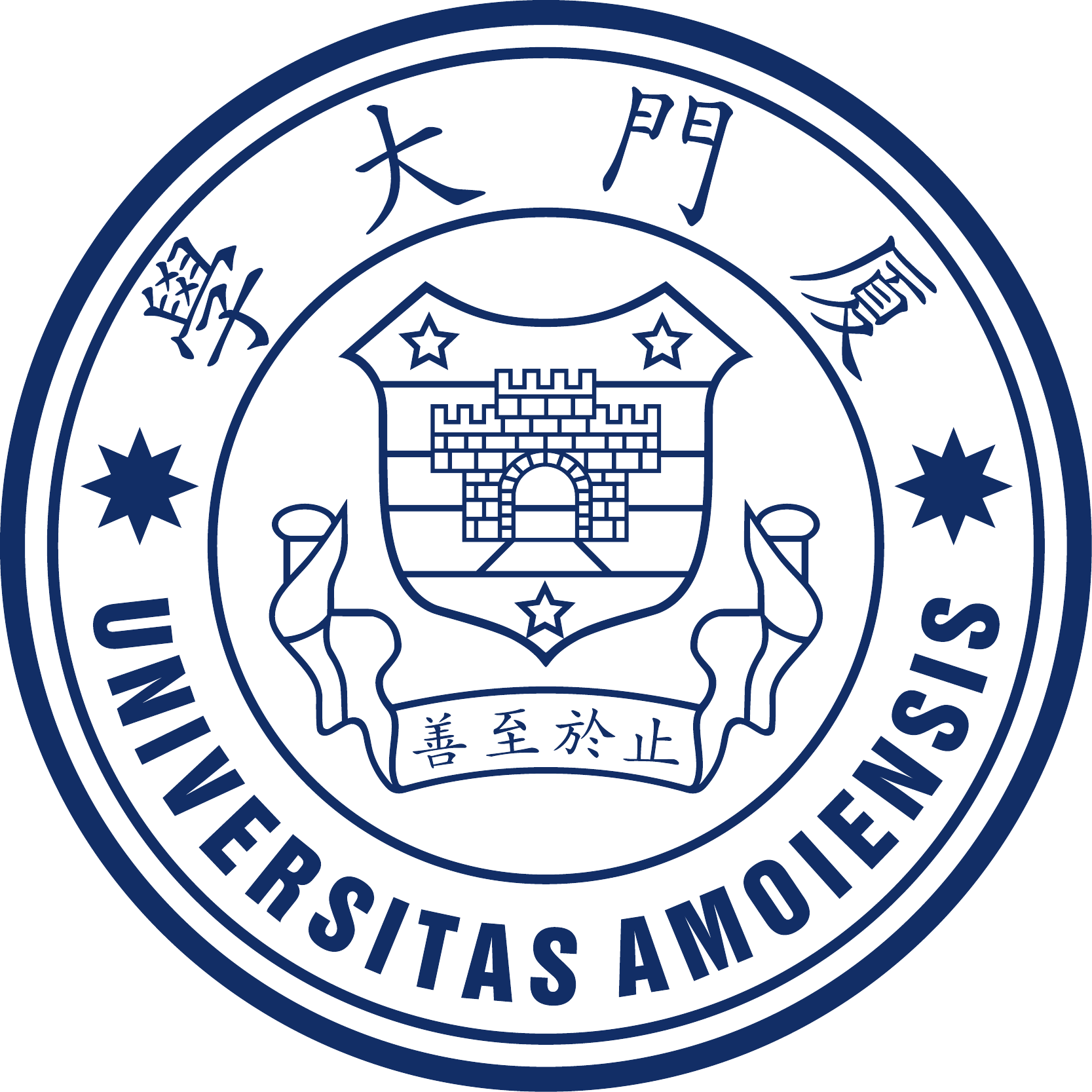}}{Xiamen University}\par}
\vspace{3.5pt}
{\noindent\fontsize{9}{11}\selectfont\color{quiet}%
\aff{\textdagger}Corresponding authors\enspace
\href{mailto:qctan@suda.edu.cn}{\nolinkurl{qctan@suda.edu.cn}} (Qichen Tan)\quad
\href{mailto:lizhijun_os@hit.edu.cn}{\nolinkurl{lizhijun_os@hit.edu.cn}} (Zhijun Li)\par}
\vspace{8pt}
\begin{tcolorbox}[width=\linewidth,colframe=abstractfill,colback=abstractfill,
  boxrule=0pt,arc=6pt,outer arc=6pt,boxsep=0pt,left=13pt,right=13pt,
  top=9pt,bottom=9pt,before skip=0pt,after skip=0pt]
{\coversans\fontseries{sb}\fontsize{8.1}{10}\selectfont\color{quiet}\textls[130]{ABSTRACT}\par}
\vspace{4pt}
{\noindent\color{coverink}\fontsize{10.1}{12.3}\selectfont Visual token merging accelerates vision--language--action (VLA) policies but can reduce the attention allocated to retained visual information.
A representative of multiple patches contributes only one attention key, reducing the region's attention weight relative to unchanged nonvisual tokens.
We identify this effect as \emph{readout dilution} and isolate its impact on closed-loop control in $\pi_{0.5}$ and OpenVLA-OFT.
Reducing visual attention while retaining all tokens roughly halves LIBERO success.
Conversely, adding the log member count to uniformly merged tokens' attention logits raises $\pi_{0.5}$ success from 12.0\% to 67.5\% at 16 tokens per camera, without changing the merged representations.
An attention-mass decomposition separates a query-independent count deficit from a representation-dependent residual.
Experiments further show that encoder-only correction cannot substitute for correction along the policy's readout paths, where visual and nonvisual keys compete.
These findings motivate MOSAIC-VLA, a training-free compressor that combines a parameter-free count bias across all visual readout paths with an adaptive spatial partition to reduce merging distortion.
With 75\% fewer visual tokens, MOSAIC-VLA keeps average LIBERO success within 0.8 percentage points of the uncompressed policy in both families, with a $1.7$--$1.8\times$ request-level speedup.
Across two real-robot tasks with 100 trials per configuration in total, uniform merging succeeds once, count correction raises this to 30 successes, and the full method reaches 85, compared with 81 for the uncompressed policy.
\par}
\vspace{5pt}
{\noindent\fontsize{8.7}{10.5}\selectfont\color{quiet}%
\begin{tabular}{@{}>{\coversans\fontseries{sb}\selectfont}l@{\hspace{.65em}}l@{}}
Date:         & November 30, 2026\\
Code:         & \url{https://github.com/CosmicBulb/mosaic-vla}\enspace (release pending)\\
Data:         & \url{https://huggingface.co/datasets/CosmicBulb/MOSAIC-VLA-evaluation}\enspace (release pending)\\
Project page: & \url{https://cosmicbulb.github.io/MOSAIC-VLA-project/}
\end{tabular}\par}
\end{tcolorbox}
\vspace{12pt}
{\captionsetup{font=covercaption,margin=6pt,labelfont={sf,bf,color=coverink}}{\centering\captionsetup{hypcap=false,skip=5pt}
\includegraphics[width=\linewidth]{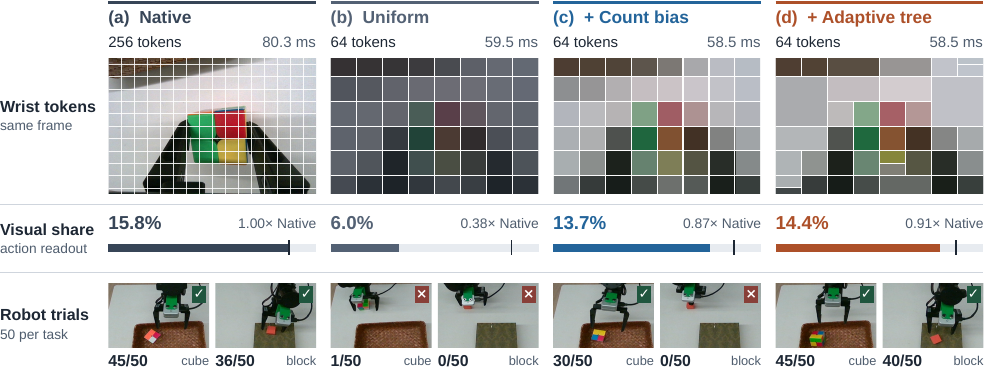}
\captionof{figure}{\textbf{Count correction and adaptive partition for visual token merging.} Four configurations of a frozen $\pi_{0.5}$ policy on an R1 Lite arm, where (b)--(d) merge 256 visual tokens per camera into 64. \textbf{Wrist tokens:} one saved cube-task frame rendered by region RGB means (a schematic of feature pooling, dimmed by visual share relative to Native) with cube-task median request latency. Columns (b) and (c) take identical tokens pooled from $2\times2$ patch regions and differ only in the $+\log 4$ bias, whereas (d) partitions regions adaptively and applies $+\log|G|$. \textbf{Visual share:} action-path visual attention share averaged across both tasks (200 replayed inputs per task, black tick marks Native). \textbf{Robot trials:} final frames from one paired block per task matching its most frequent outcome pattern, and closed-loop successes over 50 paired blocks per task.}
\label{fig:overview}
\par}
}
\clearpage

\section{Introduction}
Vision--language--action (VLA) policies now perform many manipulation tasks from camera images and a language instruction~\cite{openvla,oft,pi05}.
At each policy query, every camera view contributes hundreds of visual tokens, which dominate the policy's input sequence. Processing these tokens limits how often the policy can incorporate new observations.
We ask what a frozen VLA loses when these tokens are merged, and how much of that loss is avoidable without training.

Existing VLA compressors focus on what visual content to preserve, either by pruning tokens with low attention or action relevance~\cite{adp,safe,vlapruner} or by merging visual tokens into fewer representatives~\cite{team,depthcache,stmerge}.
At tight budgets, however, compression can sharply reduce closed-loop success.
Preserving visual representations is only part of the problem, because merging also changes how visual tokens compete against nonvisual tokens in attention.
Can this shift in cross-modal readout impair closed-loop control on its own?

A merged token carries a pooled representation of several patches but contributes only one key to attention (Fig.~\ref{fig:overview}).
When $s_G$ patches and their representative have the same attention logit, merging reduces the region's unnormalized attention mass to $1/s_G$ of its original value. With nonvisual logits unchanged, the visual attention share falls.
We call this \emph{readout dilution} and isolate it with two closed-loop interventions on LIBERO.
Without removing any visual token, scaling the action path's visual attention share to about one quarter of its native value, matching the measured mean relative share under uncorrected uniform merging at 16 tokens per camera, lowers $\pi_{0.5}$ success from 99.0\% to 53.0\% (Fig.~\ref{fig:law}c).
Conversely, holding uniformly merged tokens fixed at 16 tokens per camera and adding each token's log member count to its attention logit raises $\pi_{0.5}$ success from 12.0\% to 67.5\%.
OpenVLA-OFT shows both effects as well. These interventions show that changing visual readout weights can alter closed-loop success with the input visual representations held fixed.

An attention-mass decomposition explains this deficit. Its query-independent count term $\log s_G$ predicts the measured drop in visual log-odds across budgets, while the remaining residual depends on the representation.
With adaptive tree-based merging inside $\pi_{0.5}$'s vision encoder at 16 tokens per camera, applying the count bias only in vision-encoder layers, as encoder-side token merging does~\cite{tome}, achieves 31.5\% success versus 30.5\% uncorrected, whereas applying it on the policy's readout paths, where visual and nonvisual keys compete, raises success to 74.5\%.
Size-weighted attention itself is known from vision transformers and multimodal models~\cite{tome,pact}. For closed-loop VLA policies, we measure recovered control, pinpoint where action generation needs it, and note what it cannot recover: intra-region distinctions lost to averaging.

\begin{figure}[htbp]
\centering
\captionsetup{skip=4pt}
\includegraphics[width=\textwidth]{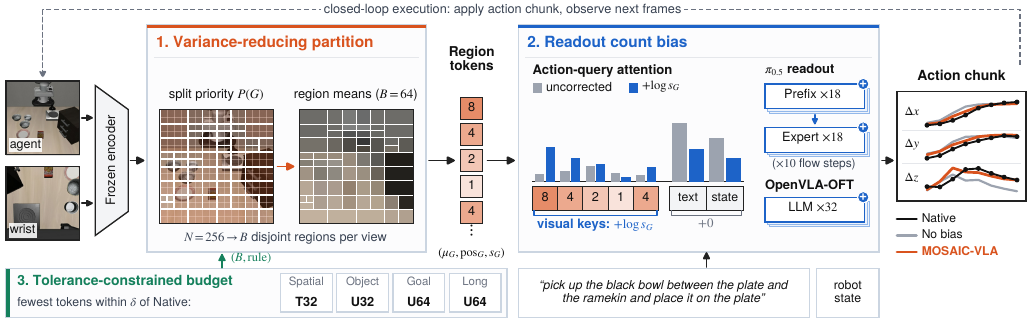}
\caption{\textbf{MOSAIC-VLA in the control loop.} \textbf{Online:} (1) The variance-reducing partition splits each camera grid by look-ahead priority $P(G)$ and pools region features into tokens $(\mu_G,\mathrm{pos}_G,s_G)$ (recorded OpenVLA-OFT Tree partition at $B=64$, pooling shown by region RGB means). (2) The readout count bias adds $\log s_G$ to attention logits for merged visual keys in every readout layer marked $+$, with zero bias on nonvisual keys (attention bars are schematic). \textbf{Right:} predicted translation commands ($\Delta x,\Delta y,\Delta z$) on the same OpenVLA-OFT request for Native, Tree without the bias (``No bias''), and Tree with the bias (``MOSAIC-VLA''). \textbf{Offline:} (3) The tolerance-constrained budget selects per-suite $(B,\text{rule})$ on development data for tolerance $\delta$ ($\pi_{0.5}$ choices at $\delta=5$ points, T: Tree, U: Uniform).}
\label{fig:method}
\end{figure}

Guided by this analysis, we propose MOSAIC-VLA, a training-free compressor for frozen VLAs that pairs readout count correction with a content-adaptive partition (Fig.~\ref{fig:method}).
A \emph{readout count bias} adds $\log s_G$ to every merged token's attention logit wherever the policy weighs visual against nonvisual tokens. At an attention layer, this correction exactly preserves the output when each representative shares its members' keys and values and the other tokens remain unchanged. It has no tunable parameter and transfers to other merging rules.
A \emph{variance-reducing partition} (Tree) divides each camera view along a binary spatial hierarchy to keep the representation residual small.
Offline, a \emph{tolerance-constrained budget} selects the fewest tokens per task suite whose development success stays within a prescribed tolerance of the uncompressed policy (Native).

We evaluate MOSAIC-VLA on more than 80,000 LIBERO and LIBERO-Plus episodes with $\pi_{0.5}$ and OpenVLA-OFT, and on two tasks with an R1 Lite robot. In simulation, count correction improves closed-loop success across 16, 32, and 64 tokens per camera, while adaptive partitioning adds further gains at tighter budgets and under LIBERO-Plus perturbations. Real-robot trials further demonstrate their complementary roles: across 100 trials per configuration, with the compressed configurations at 64 tokens per camera, uniform merging succeeds once, adding the count bias raises this to 30, and combining it with adaptive partitioning reaches 85, compared with 81 for Native (Fig.~\ref{fig:overview}).
Our contributions are:
\begin{itemize}
\item We identify readout dilution as a source
of closed-loop degradation in compressed VLA policies.
Controlled interventions establish its impact in two
policy families, while an attention-mass decomposition
explains the missing count term and experiments identify
the readout paths requiring correction.

\item We propose MOSAIC-VLA, a training-free
compressor that combines a parameter-free readout count
bias with an adaptive spatial partition, together with
a development-based rule for selecting deployment budgets.

\item We evaluate MOSAIC-VLA on LIBERO, LIBERO-Plus, and two real-robot tasks. Using 75\% fewer visual tokens, it matches Native within 0.8 LIBERO success points while delivering a $1.7$--$1.8\times$ request-level speedup. At equal budgets, Tree with count correction outperforms three public VLA pruners on OpenVLA-OFT.
\end{itemize}

\section{Related Work}

\paragraph{Efficient VLA Inference.}
Prior work allocates computation adaptively~\cite{fastvla,spvla,actionjnd,rolesub}, adapts the executed action chunk~\cite{aac}, reuses or speculates over visual representations~\cite{vlacache,efficientvla,specprunevla}, skips blocks of the action diffusion transformer~\cite{ttsparsity}, or quantizes the backbone and action head after training~\cite{quantvla}. Other methods learn pruning, compression, or active visual attention~\cite{lightvla,compressorvla,avavla}.
MOSAIC-VLA requires no training and shortens the visual sequence per query with policy weights and action settings fixed.

\paragraph{Visual Token Compression.}
Multimodal language models reduce visual tokens by attention ranking, context-based merging, and learned adaptive pruning~\cite{fastv,visionzip,atpllava}, and vision transformers do so by reconstruction-based pooling and local-then-global merging~\cite{tokenpool,algm}. For VLA policies, ADP, SAFE-Pruner, and VLA-Pruner prune visual tokens using action dynamics, attention forecasts, or action relevance~\cite{adp,safe,vlapruner}. Other compressors study spatial coverage~\cite{geoscaffold} or average visual neighborhoods~\cite{team,depthcache,stmerge}.
These VLA compressors focus on selecting or pooling visual tokens. The cited VLA merging methods do not report count correction where visual and nonvisual keys compete.

\paragraph{Size-Weighted Attention and Partitions.}
Token merging (ToMe) adds the logarithm of each representative's size to its attention logit inside the vision transformer~\cite{tome}. Later methods use saliency-weighted sizes in video transformers or weight attention by size in the vision encoder~\cite{vidtldr,dymu,tosa}, and PACT applies proportional attention after merging in an early language-model layer~\cite{pact}. The size term has also been modulated by positional distance~\cite{restore}, and message error has been bounded by a representation term plus an attention-mass mismatch~\cite{gmc}.
Analyses of how multimodal models read vision locate visual heads, find many visual tokens dispensable, and amplify visual attention against hallucination~\cite{whatsintheimage,liftingveil,visualheads,embedlens,worsethanrandom,clearsight}. In VLAs, whole attention paths have been knocked out~\cite{vlatrace}, and the action expert has been given several visual depths to read~\cite{neovla}.
Hierarchical and mixed-resolution partitions adapt granularity to content~\cite{quadformer,sttm,feng2026visionweaveweavingelasticvisual}.
MOSAIC-VLA applies count correction across the policy's readout paths, including $\pi_{0.5}$'s action expert at every flow step, and derives its partition objective from the residual left after count correction.

\section{Understanding Readout Dilution}

\begin{figure}[htbp]
\centering
\includegraphics[width=\textwidth]{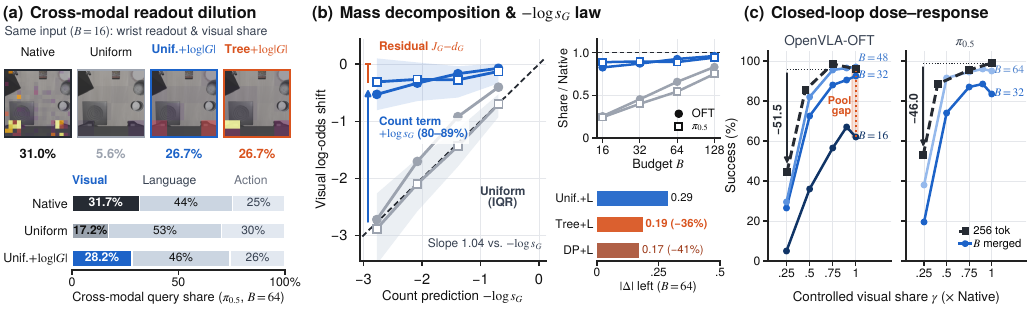}
\caption{\textbf{Readout dilution, attention-mass decomposition, and closed-loop response.} (a) Top: wrist-camera readout density and total visual share on one LIBERO input at $B=16$. Bottom: cross-modal attention distribution across 200 inputs in $\pi_{0.5}$ at $B=64$, showing nonvisual tokens absorbing lost visual share until $+\log s_G$ is applied. (b) Left: median visual log-odds shift vs.\ $-\log s_G$ across 200 inputs per model (shaded band: $\pi_{0.5}$ cell IQR), separating the count term from the residual $J_G-d_G$. Right: mean visual share normalized by Native, and residual $|\Delta|$ reduction at $B=64$ under Tree and fixed-tree DP. (c) Closed-loop success under oracle share control ($\gamma\times$ same-input Native, 200 conditions per point) for all 256 tokens (charcoal) and uniformly merged tokens at budget $B$ (blue shades).}
\label{fig:law}
\end{figure}

\label{sec:dilution}
\label{sec:problem}

\subsection{Problem Formulation}
A VLA policy $\pi$ maps camera images, an instruction, and the robot state to an action chunk. Each camera yields $N=256$ projected visual tokens $\{z_i\}$ arranged on a $16\times16$ grid, which the policy reads together with its nonvisual tokens.
A compressor $C_B$ replaces each grid by $B$ region representatives and keeps every nonvisual token. For region $G$ with $s_G=|G|$ members, it carries the mean feature, top-left original position, and member count:
\begin{equation}
r_G=(\mu_G,\mathrm{pos}_G,s_G),\qquad \mu_G=s_G^{-1}\textstyle\sum_{i\in G}z_i .
\end{equation}
Measuring success $S$ in closed loop, we seek the smallest budget whose success drops by at most $\delta$ from Native:
\begin{equation}
\min_{C_B}\ B\quad\text{s.t.}\quad S(\pi\circ C_B)\ \geq\ S(\pi)-\delta .
\label{eq:goal}
\end{equation}
MOSAIC-VLA addresses Eq.~\eqref{eq:goal} with two online steps that keep $S(\pi\circ C_B)$ high at a given $B$ and one offline rule that selects $B$ (Fig.~\ref{fig:method}). We write Native for the uncompressed policy, Uniform for merging equal regions, Tree for our variance-reducing partition, and \corr{} for adding the count bias $\log s_G$ ($s_G=|G|$). Uniform\corr{} and Tree\corr{} are the two MOSAIC-VLA configurations.

\subsection{Readout Dilution in Closed Loop}
\label{sec:surface}
We test three readout interventions on 200 paired LIBERO conditions, keeping visual-token construction fixed within each comparison (protocol in Sec.~\ref{sec:experiments}).
\runin{All Tokens Kept, Diluted Readout.} Keeping all 256 tokens per camera, we scale only the action path's visual share (the attention that action queries place on visual keys) to 0.256 and 0.242 times same-input Native, matching the measured mean relative share of uncorrected Uniform at $B=16$. Success drops from 96.0\% to 44.5\% in OpenVLA-OFT and from 99.0\% to 53.0\% in $\pi_{0.5}$ (paired 95\% success-change intervals: $[-61.5,-41.5]$ and $[-53.5,-38.0]$, Fig.~\ref{fig:law}c).
\runin{Same Tokens, Controlled Share.} Steering the visual share of OpenVLA-OFT's $B=16$ Uniform tokens to $\gamma$ times same-input Native moves success from 5.0\% at $\gamma=0.25$ to 67.0\% at $\gamma=0.9$ (Fig.~\ref{fig:law}c). Both share-control diagnostics require a same-input Native pass and serve only as diagnostics.
\runin{Same Tokens, Count Bias.} At $B=16$, Uniform merging leaves OpenVLA-OFT and $\pi_{0.5}$ at 6.5\% and 12.0\%. Adding each merged token's log member count to its attention logit, with tokens and positions unchanged, raises success to 47.0\% and 67.5\% (Table~\ref{tab:transfer}). A second execution seed reproduces the $\pi_{0.5}$ gain (11.0\% to 71.0\%).

\subsection{Decomposing the Attention-Mass Deficit}
\label{sec:law}
\label{sec:readless}
An identity for the attention mass of a merged region explains these results. For a query $q$ in any attention layer that reads the visual tokens, let $\ell_i$ be its logits on the members of $G$, $\bar\ell_G$ their mean, and $Z_G=\sum_{i\in G}e^{\ell_i}$ their unnormalized attention mass. Then
\begin{equation}
\log Z_G=\log s_G+\bar\ell_G+J_G,\ J_G=\log\frac{1}{s_G}\sum_{i\in G}e^{\ell_i-\bar\ell_G},
\label{eq:law}
\end{equation}
with $J_G\geq0$ by Jensen's inequality. If the representative's own logit is $\tilde\ell_G=\bar\ell_G+d_G$, merging changes the region's log-mass by
\begin{equation}
\log Z_G-\tilde\ell_G=\log s_G+J_G-d_G .
\label{eq:law_merge}
\end{equation}
The identity holds on actual logits in every layer. Matching the region's unmerged mass for query $q$ requires the bias $b^\star_G(q)=\log s_G+J_G(q)-d_G(q)$. Once the partition is formed, $\log s_G$ is known independently of the query. Here $J_G(q)$ is a nonnegative logit-dispersion term, while $d_G(q)$ is the signed offset of the representative's logit from the member mean, and their difference $J_G(q)-d_G(q)$ is the representation-dependent residual. A query's visual log-odds is the log ratio of its visual to nonvisual attention mass. For a fixed query whose nonvisual logits are unchanged, summing over regions gives the shift in visual log-odds,
\begin{equation}
\Delta=\log\sum_G w_G e^{-(\log s_G+J_G-d_G)},\;\; w_G=\frac{Z_G}{\sum_H Z_H},
\label{eq:logodds}
\end{equation}
so for groups of size $s$ with $J_G=d_G=0$, visual log-odds shift by $-\log s$.
Across budgets, the count-only prediction tracks the visual log-odds shifts measured in full policy forward passes on 200 saved observations per model (Fig.~\ref{fig:law}a,b). (i) Without the bias, visual log-odds fall with the count: in $\pi_{0.5}$, the median shifts at $B=16,32,64,128$ ($-2.88$, $-2.10$, $-1.44$, $-0.70$) follow $-\log s_G$ with fitted slope 1.04. (ii) Adding the count bias removes 81--89\% of this log-odds drop in $\pi_{0.5}$ and 80--82\% in OpenVLA-OFT across the tested budgets. (iii) At $B=64$, Tree yields a smaller corrected shift magnitude than Uniform ($-0.149$ vs.\ $-0.168$ in OpenVLA-OFT), and in $\pi_{0.5}$ Tree and fixed-tree DP reduce this shift magnitude by 36\% and 41\%.

This decomposition motivates two complementary designs: restoring the attention count term and reducing within-region feature variation to mitigate representation-dependent distortion. MOSAIC-VLA implements them through readout count correction and adaptive spatial partitioning.

\section{MOSAIC-VLA}
\label{sec:method}
\subsection{Readout Count Bias}

\label{sec:count}
Each representative receives the count bias on its attention logit, while every nonvisual key receives zero bias:
\begin{equation}
\ell_{qG}=q^\top k_G/\sqrt{d_k}+\log s_G .
\label{eq:area}
\end{equation}
\par\smallskip\noindent\refstepcounter{proposition}\label{prop:exact}\textbf{Proposition~\theproposition\ (Exactness).} \textit{For any fixed query with all visual keys visible, if within every merged region $G$ the members and representative share one key and one value, and all nonvisual keys, values, and visibility are unchanged, Eq.~\eqref{eq:area} yields the same attention output and nonvisual attention weights as the uncompressed sequence.}\par\smallskip
\noindent The biased representative reproduces both the region's unnormalized mass $s_Ge^{\ell_G}$ and its value-weighted contribution $s_Ge^{\ell_G}v_G$.
The count bias is thus the query-independent part of $b^\star_G$, removing the count term $\log s_G$ and leaving the representation-dependent residual.

The bias acts where the policy reads its visual tokens. We call an attention layer a readout layer when its queries weigh visual against nonvisual keys, where the key-count loss shifts attention away from vision. Inside the vision encoder all keys are visual, so a bias there only reweights visual regions among themselves without supplying the missing $\log s_G$ term in downstream readout layers.
OpenVLA-OFT reads its visual tokens in all 32 language layers. $\pi_{0.5}$ reads them along two readout paths: the bidirectional prefix attention of its language model and its action expert's attention over the cached prefix and action tokens at every flow step.
The bias enters every readout layer as one additive mask over the key axis ($\log s_G$ on visual keys and $0$ elsewhere), with policy weights untouched.

\subsection{Variance-Reducing Partition}
\begin{table}[htbp]
\centering\small
\caption{\textbf{Main results.} (a) Full LIBERO benchmark at $B=64$ (500 episodes per suite). Tok.\ is visual tokens per camera, and Rel.\ is average success normalized by Native. Shaded rows are MOSAIC-VLA, and Uniform has no bias. (b) OpenVLA-OFT vs.\ public VLA pruners under fixed per-query token budgets on 200 shared conditions (highest success in bold). Latency is mean replay request time.}
\label{tab:main}
{\textbf{(a) Full benchmark at $B=64$}\par\smallskip}
\sbox0{\setlength{\tabcolsep}{6pt}
\begin{tabular}{lrrrrrrrrrr}\toprule
Method & Tok. & Spa. & Obj. & Goal & Long & Avg. & Rel. & ms & Speedup & TFLOPs\\\midrule
\multicolumn{11}{l}{\textit{OpenVLA-OFT}}\\
Native & 256 & 97.6 & 97.6 & 98.6 & 95.2 & 97.25 & 100.0\% & 85.99 & 1.00$\times$ & 8.94\\
Uniform & 64 & 95.4 & 97.8 & 92.2 & 82.8 & 92.05 & 94.7\% & 48.07 & 1.79$\times$ & 3.70\\
\rowcolor{oursrow}Uniform\corr{} (ours) & 64 & 97.2 & 99.0 & 97.0 & 93.4 & 96.65 & 99.4\% & 46.93 & 1.83$\times$ & 3.70\\
\rowcolor{oursrow}Tree\corr{} (ours) & 64 & 97.4 & 97.6 & 96.8 & 94.0 & 96.45 & 99.2\% & 48.15 & 1.79$\times$ & 3.70\\
\midrule
\multicolumn{11}{l}{\textit{$\pi_{0.5}$}}\\
Native & 256 & 99.0 & 98.4 & 98.0 & 92.0 & 96.85 & 100.0\% & 79.97 & 1.00$\times$ & 4.75\\
Uniform & 64 & 97.4 & 97.0 & 95.4 & 89.4 & 94.80 & 97.9\% & 47.23 & 1.69$\times$ & 1.49\\
\rowcolor{oursrow}Uniform\corr{} (ours) & 64 & 98.2 & 97.0 & 97.8 & 93.0 & 96.50 & 99.6\% & 47.41 & 1.69$\times$ & 1.49\\
\rowcolor{oursrow}Tree\corr{} (ours) & 64 & 98.4 & 97.4 & 98.6 & 92.8 & 96.80 & 99.9\% & 47.25 & 1.69$\times$ & 1.49\\
\bottomrule
\end{tabular}
}%
\setlength{\tabcolsep}{\dimexpr 6pt+(0.94\textwidth-\wd0)/22\relax}%

\begin{tabular}{lrrrrrrrrrr}\toprule
Method & Tok. & Spa. & Obj. & Goal & Long & Avg. & Rel. & ms & Speedup & TFLOPs\\\midrule
\multicolumn{11}{l}{\textit{OpenVLA-OFT}}\\
Native & 256 & 97.6 & 97.6 & 98.6 & 95.2 & 97.25 & 100.0\% & 85.99 & 1.00$\times$ & 8.94\\
Uniform & 64 & 95.4 & 97.8 & 92.2 & 82.8 & 92.05 & 94.7\% & 48.07 & 1.79$\times$ & 3.70\\
\rowcolor{oursrow}Uniform\corr{} (ours) & 64 & 97.2 & 99.0 & 97.0 & 93.4 & 96.65 & 99.4\% & 46.93 & 1.83$\times$ & 3.70\\
\rowcolor{oursrow}Tree\corr{} (ours) & 64 & 97.4 & 97.6 & 96.8 & 94.0 & 96.45 & 99.2\% & 48.15 & 1.79$\times$ & 3.70\\
\midrule
\multicolumn{11}{l}{\textit{$\pi_{0.5}$}}\\
Native & 256 & 99.0 & 98.4 & 98.0 & 92.0 & 96.85 & 100.0\% & 79.97 & 1.00$\times$ & 4.75\\
Uniform & 64 & 97.4 & 97.0 & 95.4 & 89.4 & 94.80 & 97.9\% & 47.23 & 1.69$\times$ & 1.49\\
\rowcolor{oursrow}Uniform\corr{} (ours) & 64 & 98.2 & 97.0 & 97.8 & 93.0 & 96.50 & 99.6\% & 47.41 & 1.69$\times$ & 1.49\\
\rowcolor{oursrow}Tree\corr{} (ours) & 64 & 98.4 & 97.4 & 98.6 & 92.8 & 96.80 & 99.9\% & 47.25 & 1.69$\times$ & 1.49\\
\bottomrule
\end{tabular}

\par\smallskip
{\textbf{(b) Comparison with public VLA pruners, OpenVLA-OFT}\par\smallskip}
\sbox0{\setlength{\tabcolsep}{6pt}
\begin{tabular}{l rrrr rrrr}\toprule
& \multicolumn{4}{c}{64 tokens in total, $B=32$ per camera} & \multicolumn{4}{c}{128 tokens in total, $B=64$ per camera}\\
\cmidrule(lr){2-5}\cmidrule(lr){6-9}
Method & Success & ms & Speedup & TFLOPs & Success & ms & Speedup & TFLOPs\\\midrule
Native & \multicolumn{8}{c}{512 tokens: 98.0\% at 85.99\,ms, 8.94 TFLOPs}\\
ADP (ICLR'26) & 46.5 & 45.98 & 1.87$\times$ & 2.86 & 64.5 & 46.82 & 1.84$\times$ & 3.72\\
SAFE-Pruner (ECCV'26) & 94.0 & 57.79 & 1.49$\times$ & 4.12 & 97.0 & 59.17 & 1.45$\times$ & 4.90\\
VLA-Pruner (arXiv'25) & 93.5 & 54.08 & 1.59$\times$ & 3.59 & 96.0 & 56.59 & 1.52$\times$ & 4.35\\
\rowcolor{oursrow}Uniform\corr{} (ours) & 87.0 & 45.67 & 1.88$\times$ & 2.85 & 97.0 & 46.93 & 1.83$\times$ & 3.70\\
\rowcolor{oursrow}Tree\corr{} (ours) & \textbf{95.0} & 46.44 & 1.85$\times$ & 2.85 & \textbf{97.5} & 48.15 & 1.79$\times$ & 3.70\\
\bottomrule
\end{tabular}
}%
\setlength{\tabcolsep}{\dimexpr 6pt+(0.94\textwidth-\wd0)/18\relax}%

\begin{tabular}{l rrrr rrrr}\toprule
& \multicolumn{4}{c}{64 tokens in total, $B=32$ per camera} & \multicolumn{4}{c}{128 tokens in total, $B=64$ per camera}\\
\cmidrule(lr){2-5}\cmidrule(lr){6-9}
Method & Success & ms & Speedup & TFLOPs & Success & ms & Speedup & TFLOPs\\\midrule
Native & \multicolumn{8}{c}{512 tokens: 98.0\% at 85.99\,ms, 8.94 TFLOPs}\\
ADP (ICLR'26) & 46.5 & 45.98 & 1.87$\times$ & 2.86 & 64.5 & 46.82 & 1.84$\times$ & 3.72\\
SAFE-Pruner (ECCV'26) & 94.0 & 57.79 & 1.49$\times$ & 4.12 & 97.0 & 59.17 & 1.45$\times$ & 4.90\\
VLA-Pruner (arXiv'25) & 93.5 & 54.08 & 1.59$\times$ & 3.59 & 96.0 & 56.59 & 1.52$\times$ & 4.35\\
\rowcolor{oursrow}Uniform\corr{} (ours) & 87.0 & 45.67 & 1.88$\times$ & 2.85 & 97.0 & 46.93 & 1.83$\times$ & 3.70\\
\rowcolor{oursrow}Tree\corr{} (ours) & \textbf{95.0} & 46.44 & 1.85$\times$ & 2.85 & \textbf{97.5} & 48.15 & 1.79$\times$ & 3.70\\
\bottomrule
\end{tabular}

\end{table}

\label{sec:partition}
\noindent\refstepcounter{proposition}\label{prop:bound}\textbf{Proposition~\theproposition\ (Residual bound).} \textit{If $k_i=W_Kz_i$ and $k_G=W_K\mu_G$, then $d_G=0$ and every query $q$ obeys $0\leq J_G(q)\leq\|W_K^\top q\|_2\max_{i\in G}\|z_i-\mu_G\|_2/\sqrt{d_k}$.}\par\smallskip
\noindent To second order, the query average of $J_G$ is a quadratic form in the deviations $z_i-\mu_G$. When regions are read in proportion to their size and this form is isotropic, the mass-weighted $J_G$ term is proportional to the total distortion $\sum_GE(G)$, with $E(G)=\sum_{i\in G}\|z_i-\mu_G\|_2^2$. Each representative retains its top-left patch's original position rather than taking a consecutive compressed position.

We use total within-region distortion as a second-order proxy objective. Let $\mathcal T$ be the binary spatial hierarchy that recursively bisects each region along its longer side, and let a frontier $\mathcal F\subset\mathcal T$ cover every patch once:
\begin{equation}
\min_{\mathcal F\subset\mathcal T,\;|\mathcal F|=B}\ \textstyle\sum_{G\in\mathcal F}E(G).
\label{eq:objective}
\end{equation}
Splitting $G$ into $G_L,G_R$ cuts $E$ by Ward's gain~\cite{ward},
\begin{equation}
\Delta E(G)=\frac{|G_L||G_R|}{|G_L|+|G_R|}\|\mu_{G_L}-\mu_{G_R}\|_2^2\geq0,
\label{eq:gain}
\end{equation}
and Tree ranks splits by look-ahead priority ($P=0$ at leaves)
\begin{equation}
P(G)=\Delta E(G)+\max\{P(G_L),P(G_R)\}.
\label{eq:priority}
\end{equation}
Tree selects the top $B-1$ splits. Because $P$ never increases from parent to child, breaking ties ancestor-first keeps the selected set closed under ancestors and always yields a valid frontier. Node statistics for all $2N-1$ nodes are computed bottom-up in $O(ND)$ for $D$-dimensional features, and ranking takes $O(N\log N)$. For power-of-two budgets $B$, Uniform is the equal-depth frontier of $\mathcal T$, and Tree retains Uniform whenever its candidate frontier does not lower total distortion, so Tree never exceeds Uniform's objective. Fixed-tree dynamic programming solves Eq.~\eqref{eq:objective} exactly and serves as a reference.\par\noindent\textit{Details are available in the Supplementary Material.}

\subsection{Tolerance-Constrained Budget}
\label{sec:target}
The budget rule targets Eq.~\eqref{eq:goal} on development data over a finite candidate set. Given a tolerance of $\delta$ percentage points, it selects for each task suite the configuration with the fewest visual tokens among Uniform\corr{} and Tree\corr{} at $B\in\{16,32,64,128\}$ whose development success is at most $\delta-1$ points below Native's (leaving a 1-point guard band), falling back to Native if none qualifies. Ties go to lower latency and then to Uniform\corr{}.
The selected budget and partition rule stay fixed in deployment, while Tree's regions still adapt to each input.

\begin{figure}[htbp]
\centering
\captionsetup{skip=4pt}
\includegraphics[width=\textwidth]{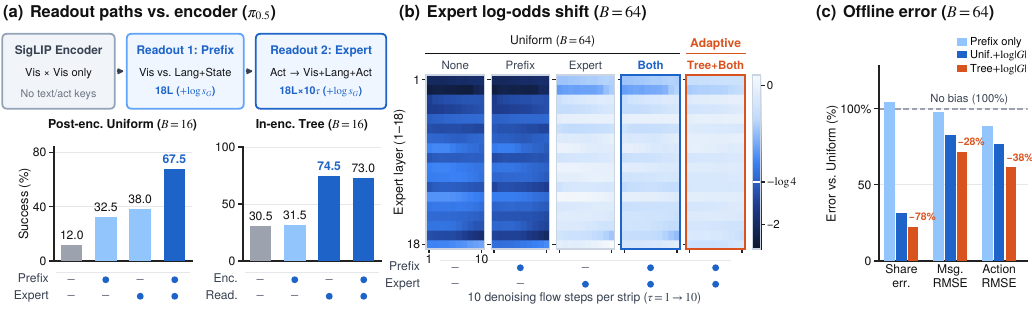}
\caption{\textbf{Dual readout paths and bias placement in $\pi_{0.5}$.} Filled dots mark where $+\log s_G$ enters. (a) Three attention stages in $\pi_{0.5}$ and closed-loop success at $B=16$ (200 paired conditions per bar) under post-encoder Uniform merging and in-encoder Tree merging. (b) Layer-by-flow-step median expert visual log-odds shifts at $B=64$ across 200 saved inputs (colorbar white tick marks $-\log 4$). (c) Offline share error, visual-message RMSE, and action RMSE at $B=64$ normalized by uncorrected Uniform.}
\label{fig:paths}
\end{figure}

\section{Experiments}
\label{sec:experiments}
\label{sec:eval}

\subsection{Setup}
\begin{figure}[htbp]
\centering
\includegraphics[width=\textwidth]{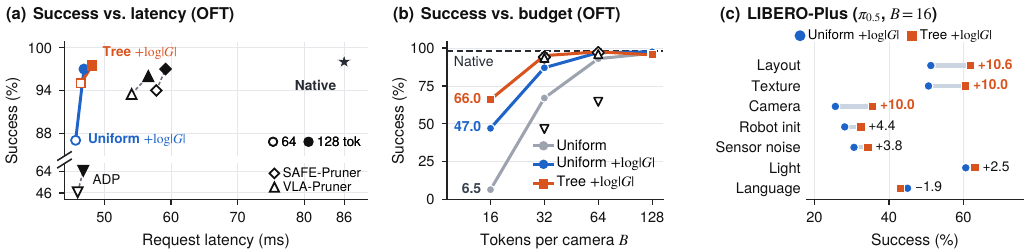}
\caption{\textbf{Success--latency trade-off and robustness.} (a) OpenVLA-OFT success on 200 shared conditions vs.\ request latency at 64 (open) and 128 (filled) total visual tokens (broken success axis). (b) OpenVLA-OFT success across per-camera budgets $B$ on the same 200 conditions for Uniform, Uniform\corr{}, and Tree\corr{} (open black markers: public pruners). (c) $\pi_{0.5}$ LIBERO-Plus success at $B=16$ by perturbation category (160 conditions per point, labels give Tree\corr{} minus Uniform\corr{} in percentage points).}
\label{fig:quality_latency_development}
\end{figure}

OpenVLA-OFT runs suite-specific checkpoints on an RTX~5090 in bfloat16 with SDPA (eight actions per request). Unless noted in Supp.\ Sec.~\ref{app:oft_tome}, $\pi_{0.5}$ runs the public LIBERO checkpoint~\cite{openpi} on an RTX~5080 (horizon 10, 5 executed actions, 10 flow steps).
Native reproduces the published averages (97.25\% vs.\ 97.1\% and 96.85\% vs.\ 96.85\%~\cite{oft,openpi}) with weights and action protocols fixed. Baselines are three VLA pruners, ADP, SAFE-Pruner, and VLA-Pruner~\cite{adp,safe,vlapruner}, evaluated from their public OpenVLA-OFT implementations under fixed per-query budgets of 64 and 128 total visual tokens without full-token fallback.
LIBERO-Spatial, Object, Goal, and Long each contain ten tasks with 50 published initial states~\cite{libero}, and a condition is a task--initial-state pair.
Full-benchmark experiments comprise 56,000 episodes across all 2,000 conditions with execution seed 7 and an additional 16,000 episodes at $B=64$ with execution seed 13, while 200-condition panels use initial states 0--4.
Intervals resample whole tasks within suites with initial states and configurations paired, with 200-condition families and confidence levels in Supp.\ Sec.~\ref{app:pi05ports}.
Within each model family, all configurations replay the same 40 recorded sequences twice, yielding 1,530 measured requests per configuration for OpenVLA-OFT and 2,518 for $\pi_{0.5}$. TFLOPs count partial dense-equivalent matrix, convolution, and attention ops.

\subsection{Main Results}
\label{sec:main}
\label{sec:baselines}
Readout correction keeps closed-loop success high with 75\% fewer visual tokens. At $B=64$ on the full benchmark (Table~\ref{tab:main}a), adding the count bias lifts Uniform from 92.05\% to 96.65\% in OpenVLA-OFT and from 94.80\% to 96.50\% in $\pi_{0.5}$, while Tree\corr{} reaches 96.45\% and 96.80\%. Both MOSAIC-VLA configurations keep average success within 0.80 percentage points of Native while serving requests 1.79--$1.83\times$ faster in OpenVLA-OFT and $1.69\times$ faster in $\pi_{0.5}$. On Uniform, the bias gains most on the long-horizon LIBERO-Long suite ($+10.6$ points in OpenVLA-OFT and $+3.6$ in $\pi_{0.5}$), and execution seed 13 reproduces both the LIBERO-Long ($+8.8$ and $+6.2$) and overall Uniform gains ($+3.85$ and $+2.80$) across all 2,000 conditions. Uncorrected failures run to the step limit, whereas with the bias mean requests per episode stay within about 2.8\% of Native. Multiplying mean requests per episode by replay latency estimates that Uniform\corr{} cuts episode inference time by 45.6\% in OpenVLA-OFT and 39.0\% in $\pi_{0.5}$ vs.\ Native.

On 200 shared OpenVLA-OFT conditions at matched total budgets of 64 and 128 visual tokens (Table~\ref{tab:main}b), Tree\corr{} achieves the highest observed success among the compared compressed configurations (95.0\% and 97.5\%). SAFE-Pruner and VLA-Pruner take 23--24\% and 16--18\% more request time than Tree\corr{}. Uniform\corr{} at 64 total tokens also achieves higher success and lower mean request latency than both ADP settings (Fig.~\ref{fig:quality_latency_development}a). Across all four tested budgets, adding the count bias raises Uniform success (Fig.~\ref{fig:quality_latency_development}b).

\begin{figure}[htbp]
\centering
\captionsetup{skip=4pt}
\includegraphics[width=\textwidth]{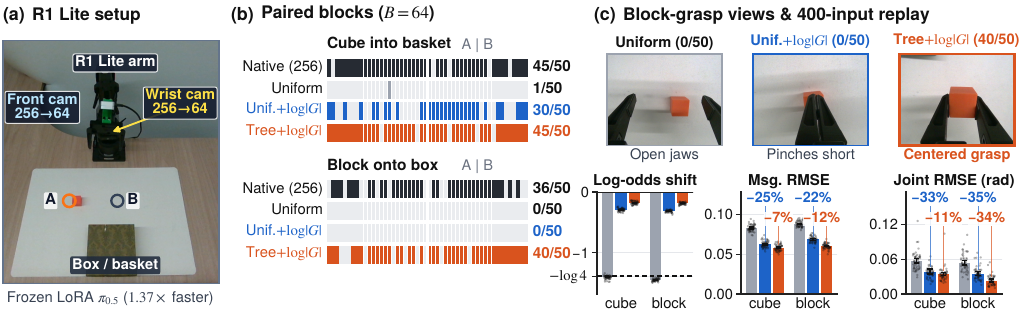}
\caption{\textbf{Real-robot setup, paired outcomes, and readout diagnostics.} (a) Dual-camera R1 Lite setup with marked start positions A/B and target zone. (b) Closed-loop outcomes across 50 paired blocks per task ($B=64$ for compressed arms, 256 for Native). (c) Top: wrist-camera grasp views on the block task comparing Uniform, Uniform\corr{}, and Tree\corr{}. Bottom: offline replay of 200 saved inputs per task (50-trial means and 95\% CIs, dots show per-trial means, and labels give stepwise percentage changes).}
\label{fig:robot}
\end{figure}

\subsection{Component Analysis}
\label{sec:plugin}
\label{sec:standard}
\begin{table}[htbp]
\centering\footnotesize
\captionsetup{skip=4pt}
\caption{\textbf{Merging rules on 200 shared LIBERO conditions} (success, \%). U/T denote Uniform/Tree, ToMe is within-camera similarity merging, and +L adds $+\log s_G$.}
\label{tab:transfer}
\setlength{\tabcolsep}{4.6pt}

\begin{tabular}{@{}lr rrrrr@{}}\toprule
Model & $B$ & U & U+L & ToMe & ToMe+L & T+L\\\midrule
OFT & 16 & 6.5 & 47.0 & 3.5 & 13.5 & \textbf{66.0}\\
OFT & 32 & 67.0 & 87.0 & 43.0 & 75.5 & \textbf{95.0}\\
OFT & 64 & 93.0 & 97.0 & 91.5 & \textbf{99.0} & 97.5\\
\midrule
$\pi_{0.5}$ & 16 & 12.0 & 67.5 & 3.0 & 25.5 & \textbf{77.0}\\
$\pi_{0.5}$ & 32 & 71.5 & 86.0 & 58.0 & 72.5 & \textbf{89.5}\\
$\pi_{0.5}$ & 64 & 94.0 & \textbf{96.0} & 91.5 & 92.0 & 95.5\\
\bottomrule
\end{tabular}

\end{table}
\textbf{Placement of the Count Bias.}
In $\pi_{0.5}$ under post-encoder Uniform merging at $B=16$, correcting only the language-model prefix, where multimodal methods apply size-weighted attention~\cite{pact}, gives 32.5\% success. Also correcting the action expert at every flow step reaches 67.5\%, 35.0 and 29.5 percentage points above correcting the prefix or the expert alone (95\% intervals $[24.5,45.0]$ and $[19.0,40.0]$, Fig.~\ref{fig:paths}a). Without correction, the expert's visual log-odds shift by about the count prediction across layers and flow steps. Correcting the expert path removes most of this shift, whereas prefix-only correction does not (Fig.~\ref{fig:paths}b). Correcting both readout paths lowers offline errors, with further reductions from Tree (Fig.~\ref{fig:paths}c).
When $\pi_{0.5}$ merges inside its vision encoder with Tree at $B=16$, biasing only the remaining encoder layers, as encoder-side merging does~\cite{tome}, leaves success near the uncorrected baseline (31.5\% vs.\ 30.5\%), whereas biasing only the readout paths reaches 74.5\% ($+44.0$ percentage points, family-adjusted $[31.0,57.0]$, Fig.~\ref{fig:paths}a).
Rescaling the bias ($\alpha\log s_G$) gives no gain in OpenVLA-OFT: at $B=32$, $\alpha=1$ reaches 87.0\% vs.\ 83.0\% and 81.5\% at $\alpha=0.5$ and $1.5$.
\runin{Transfer Across Merging Rules.}
For a similarity-based merging adapter matching projected features within each camera~\cite{tome}, the count bias raises OpenVLA-OFT success by 10.0, 32.5, and 7.5 percentage points at $B=16$, 32, and 64 and $\pi_{0.5}$ success by 14.5 points at $B=32$, with all four family-adjusted intervals above zero (Table~\ref{tab:transfer}). Merging ADP's discarded patches into kept tokens with $+\log s_G$ also lifts ADP from 46.5\%/64.5\% to 55.5\%/70.5\%.
\runin{Effect of the Partition.}
At matched token budgets with count correction applied, Tree\corr{} exceeds Uniform\corr{} by 19.0 and 8.0 percentage points in OpenVLA-OFT at $B=16$ and 32 (98.75\% intervals $[9.5,30.0]$ and $[2.0,15.0]$) and beats biased similarity merging by 52.5 and 19.5 points, while at $B=64$ the three rules lie within 2 points. In $\pi_{0.5}$ at $B=32$, Tree\corr{} leads them by 3.5 and 17.0 points (Table~\ref{tab:transfer}). Rescaling Uniform's bias at $B=16$ stays below 50\%, vs.\ 66.0\% for Tree\corr{}.
\runin{Tolerance-Constrained Budget.}
With $\delta=5$ percentage points on 800 held-out conditions per model, the rule uses 48 visual tokens per camera in $\pi_{0.5}$ and 40 in OpenVLA-OFT, 25\% and 37.5\% fewer than $B=64$, at 2.38 and 3.38 points below Native (choices in Fig.~\ref{fig:method}).

\subsection{Real Robot and Robustness}
\label{sec:target_results}
\label{sec:robust}
\label{sec:robot}
\begin{table}[htbp]
\centering\footnotesize
\captionsetup{skip=4pt}
\caption{\textbf{LIBERO-Plus robustness in $\pi_{0.5}$} across 1,120 perturbation conditions per cell (success, \%).}
\label{tab:plus}
\setlength{\tabcolsep}{4.8pt}

\begin{tabular}{@{}r r rrrr@{}}\toprule
$B$ & Native & Uniform & Unif.\corr{} & Tree & Tree\corr{}\\\midrule
16 & 85.6 & 3.4 & 41.7 & 6.9 & \textbf{47.3}\\
32 & 85.4 & 44.1 & 63.2 & 56.2 & \textbf{71.5}\\
64 & 85.5 & 75.4 & 79.7 & 78.0 & \textbf{80.5}\\
\bottomrule
\end{tabular}

\end{table}
\textbf{Real Robot.} We evaluate one LoRA-fine-tuned two-camera $\pi_{0.5}$ policy on an R1 Lite arm across two tasks, placing a cube into a basket and a block onto a box (Fig.~\ref{fig:robot}a). Each task comprises 50 paired blocks, 25 per marked start position, evaluating Native (256 tokens per camera) and three compressed configurations at $B=64$ once per block.
On the cube task, uniform merging succeeds in 1/50 trials, adding the count bias raises this to 30/50, and Tree\corr{} reaches 45/50, matching Native's 45/50. Within blocks, the two steps win their discordant pairs 29:0 and 17:2 (Holm-adjusted exact McNemar $p<0.0001$ and $p=0.002$).
On the block task, both Uniform configurations fail all 50 trials, lifting the block in only 1 of their 100 trials, whereas Tree\corr{} succeeds in 40/50 versus 36/50 for Native, winning all 40 discordant pairs against Uniform\corr{} ($p<0.0001$, Fig.~\ref{fig:robot}b,c).
Offline replay of 200 saved inputs per task through the deployed checkpoint traces the same progression from readout to action (Fig.~\ref{fig:robot}c). Uniform merging shifts action-path visual log-odds by $-1.41$ and $-1.45$, close to $-\log 4=-1.39$. Adding the count bias reduces the remaining shift to $-0.29$ and $-0.32$ and cuts arm joint-command RMSE by about a third on both tasks. Adding Tree further lowers visual-message RMSE by 7\% and 12\% and arm joint-command RMSE by 11\% and 34\% on the cube and block tasks, while achieving a $1.37\times$ request speedup on each task.
\runin{Perturbations.} On 1,120 LIBERO-Plus perturbation conditions~\cite{liberoplus} in $\pi_{0.5}$ (Table~\ref{tab:plus}), adding the count bias raises both partitions by 38--40 points at $B=16$ and 15--19 points at $B=32$. With the bias applied, Tree\corr{} leads Uniform\corr{} by 5.6 points at $B=16$ (95\% interval $[2.4,8.9]$) and 8.3 points at $B=32$, leading in six of seven categories and by 10.0--10.6 points under texture, camera, and layout shifts (Fig.~\ref{fig:quality_latency_development}c).\par\noindent\textit{Details are available in the Supplementary Material.}

\section{Conclusion}
An attention-mass decomposition separates the count deficit of visual token merging from representation distortion. MOSAIC-VLA addresses these effects with readout count correction and a variance-reducing partition, without training. With 75\% fewer visual tokens, it keeps average LIBERO success within 0.8 percentage points of Native in $\pi_{0.5}$ and OpenVLA-OFT while accelerating requests by $1.7$--$1.8\times$. Across two real-robot tasks with 100 trials per configuration, MOSAIC-VLA at 64 tokens per camera succeeds 85 times, compared with 1 for uniform merging and 81 for Native. Token budgets here are fixed per task suite, making per-observation budget adaptation and broader multi-task manipulation natural directions for future work.

\par\FloatBarrier\medskip\phantomsection\clearpage\phantomsection\addcontentsline{toc}{section}{Supplementary Material}
{\sffamily\footnotesize\color{quiet}MOSAIC-VLA\par}\vspace{3pt}
{\LARGE\bfseries\color{ink}Supplementary Material\par}\vspace{8pt}

\appendices
\begingroup
\newcommand{\mainref}[1]{\ref{#1}}
\newcommand{\maineqref}[1]{(\ref{#1})}

This supplement has four parts: theory and algorithms (Part~\ref{supp:theory}), setup, timing, and statistics (Part~\ref{supp:setup}), mechanism, ablations, and additional results (Part~\ref{supp:mechanism}), and the real-robot evaluation (Part~\ref{supp:robot}). The accompanying video plays one paired block of each real-robot task with all four configurations side by side, and the paired LIBERO episode of Fig.~\ref{fig:episode}. The list below locates the most frequently consulted material.

\par\smallskip\noindent{\small\setlength{\tabcolsep}{3pt}%
\begin{tabular}{@{}p{0.80\columnwidth}r@{}}\toprule
Material & Location\\\midrule
Readout identity and residual bound, with proofs & \ref{app:law}\\
Exact partition reference (fixed-tree DP) & \ref{app:dp}\\
Algorithms~\ref{alg:budget} and~\ref{alg:mosaic} & \ref{app:algorithms}\\
Episode accounting and reproduction & \ref{app:reproduction}\\
Request latency of every configuration & \ref{app:timing}\\
Comparison families and baseline scope & \ref{app:pi05ports}\\
Readout diagnostics on saved inputs & \ref{app:readout_metrics}--\ref{app:keytypes}\\
Bias placement and readout-route ablations & \ref{app:readout_closed_loop}--\ref{app:placement}\\
Second execution seed on the full benchmark & \ref{app:seed13}\\
Full-benchmark success by suite & \ref{app:bysuite}\\
Three-budget similarity merging (Table~\mainref{tab:transfer}) & \ref{app:oft_tome}\\
Bias-scale sweep and pruner-merge extension & \ref{app:alpha_adp}\\
LIBERO-Plus confirmation and budget extensions & \ref{app:plus_confirmation}\\
Real-robot protocol, outcomes, and raw counts & \ref{app:robot}\\
Real-robot readout diagnostics & \ref{app:r1lite}\\
Paired real-robot blocks and LIBERO episode & video\\
\bottomrule
\end{tabular}}\par\smallskip

\section{Theory and Algorithms}
\label{supp:theory}

\subsection{Readout Law}
\label{app:law}
Eq.~\maineqref{eq:law} is an identity on the logits $\ell_i$ of one query: write $x_i=\ell_i-\bar\ell_G$, so $\frac{1}{s_G}\sum_i x_i=0$ and $J_G=\log\frac{1}{s_G}\sum_i e^{x_i}$.
Jensen's inequality gives $J_G\geq0$.
A second-order expansion with remainder, $e^{x}\leq1+x+\frac12x^2e^{r_G}$ for $x\leq r_G=\max_i x_i$, together with $\log(1+y)\leq y$, gives
\begin{equation}
0\leq J_G\leq\tfrac12 e^{r_G}\operatorname{Var}_{i\in G}\ell_i,
\end{equation}
so $J_G\approx\frac12\operatorname{Var}_{i\in G}\ell_i$ for a small spread, and higher-order terms grow under strong compression.
With keys $k_i=W_Kz_i$ and the representative's key $W_K\mu_G$, the representative's logit is the mean of its members' logits, so $d_G=0$, and Cauchy--Schwarz bounds the residual for every query: $0\leq J_G(q)\leq\|W_K^\top q\|_2\max_{i\in G}\|z_i-\mu_G\|_2/\sqrt{d_k}$.
If keys are linear in features, $\ell_i=a_q^\top z_i+c_q$ with $a_q=W_K^\top q/\sqrt{d_k}$, then $\operatorname{Var}_{i\in G}\ell_i=a_q^\top\Sigma_Ga_q\leq\|a_q\|^2E(G)/s_G$ with $\Sigma_G$ the within-region covariance.
Averaging over queries gives
\begin{equation*}
\mathbb{E}_q\bigl[\operatorname{Var}_{i\in G}\ell_i\bigr]=\frac{1}{s_G}\sum_{i\in G}(z_i-\mu_G)^\top M(z_i-\mu_G)
\end{equation*}
with $M=\mathbb{E}_q[a_qa_q^\top]=W_K^\top\mathbb{E}[qq^\top]W_K/d_k$, summed over heads.
Isotropic queries give $M\propto W_K^\top W_K$ rather than the identity. Euclidean distortion is the case $M=cI$, which needs no query statistics.
For $M=cI$, $s_G\,\mathbb{E}_q[J_G]\approx\frac{c}{2}E(G)$, so the total distortion $\sum_GE(G)$ that Tree and DP minimize equals, to second order, the mass-weighted heterogeneity of Eq.~\maineqref{eq:logodds} when regions are read in proportion to their size.

Eq.~\maineqref{eq:logodds} follows from Eq.~\maineqref{eq:law_merge}: $e^{\tilde\ell_G}=Z_G\,e^{-(\log s_G+J_G-d_G)}$, and summing over regions compares the merged readout with the same query's readout of the unmerged member keys.
Writing $x_G=J_G-d_G$, Jensen's inequality gives $-\sum_Gw_Gx_G\leq\log\sum_Gw_Ge^{-x_G}\leq-\min_Gx_G$ for the corrected readout.
Across layers, upstream changes also move queries and nonvisual keys, and Fig.~\mainref{fig:law} measures the resulting shift end to end.
The offset $d_G=\tilde\ell_G-\bar\ell_G$ captures effects specific to the representative key, such as positional encoding and layer normalization.
Merging also changes the value aggregation: to first order, the attention-weighted member mean differs from $\mu_G$ by the within-group covariance of logits and values. Because $J_G$ captures attention-mass dispersion while value aggregation also affects the output, we evaluate partitions by downstream action error and closed-loop success.

\subsection{Tree-Optimal Distortion Allocation}
\label{app:dp}
The look-ahead priority of Eq.~\maineqref{eq:priority} is a greedy heuristic rather than an exact solver for fixed-budget distortion minimization.
On a fixed binary hierarchy, the minimum within-region distortion for subtree $G$ at budget $b$ satisfies the dynamic program
\begin{align}
 F_G(1)&=E(G),\\
 F_G(b)&=\min_{b_L+b_R=b}\left[F_{G_L}(b_L)+F_{G_R}(b_R)\right],
\end{align}
for feasible $b\ge2$ with positive child budgets $b_L,b_R$, yielding the exact distortion-minimizing frontier within $\mathcal T$.

In $\pi_{0.5}$, we evaluate fixed-tree DP in closed loop both with and without the count bias (Table~\ref{tab:allocations}), whereas the five-configuration OpenVLA-OFT benchmark evaluates Uniform and Tree. On CPU at $B=64$ across two views, DP frontier selection from precomputed scalar node costs takes 0.76--0.80\,ms, and full CPU execution including feature statistics and token packing takes 8.13\,ms for DP versus 4.41\,ms for Tree.

\subsection{Algorithms}
\label{app:algorithms}
Algorithm~\ref{alg:budget} builds the Tree partition of one camera view, and Algorithm~\ref{alg:mosaic} runs MOSAIC-VLA for one request.
\begin{algorithm}[tb]
\small\DontPrintSemicolon
\caption{Hierarchical spatial budgeting for one view}
\label{alg:budget}
\KwIn{Projected grid $Z$, binary hierarchy $\mathcal{T}$, budget $B$}
\KwOut{Region means, original positions, and member counts}
Compute node means and split gains $\Delta E$ bottom-up\;
Compute priorities $P$ using Eq.~\maineqref{eq:priority}\;
Sort internal nodes by $P$, ancestors before descendants in ties\;
Select the first $B-1$ splits and obtain frontier $\mathcal{F}$\;
Retain Uniform if the candidate does not reduce distortion\;
Order regions by their original representative positions\;
Pack $(\mu_G,\operatorname{pos}(G),|G|)$ for each region\;
\end{algorithm}

\begin{algorithm}[htbp]
\small\DontPrintSemicolon
\caption{MOSAIC-VLA for one request}
\label{alg:mosaic}
\KwIn{Projected grid $\{z_i\}$ of each camera, with budget $B$ and partition rule fixed per suite by the target rule}
\KwOut{Action chunk}
\ForEach{camera}{
\eIf{rule is Tree}{
compute node means $\mu_G$, sizes $s_G$, and gains $\Delta E(G)$ on $\mathcal T$ bottom-up\;
take the $B-1$ splits with the largest $P(G)$, Eq.~\maineqref{eq:priority}, ancestors first in ties, to form $\mathcal F$\;
fall back to the Uniform frontier if $\sum_{G\in\mathcal F}E(G)$ is not lower\;
}{take the Uniform frontier $\mathcal F$ and its region means\;}
emit $(\mu_G,\operatorname{pos}(G),s_G)$ for $G\in\mathcal F$ in source order, $\operatorname{pos}(G)$ being the top-left source position\;
}
run the frozen policy, all nonvisual tokens kept, with $q^\top k_G/\sqrt{d_k}+\log s_G$ on every visual key in every readout path, Eq.~\maineqref{eq:area}\;
\end{algorithm}

\section{Setup, Timing, and Statistics}
\label{supp:setup}

\subsection{Reproducibility and Cost Accounting}
\label{app:reproduction}
The OpenVLA-OFT primary and external comparisons comprise 24,000 episodes, and the $\pi_{0.5}$ full-benchmark matrix comprises 32,000 episodes across 16 configurations. Execution seed 13 adds 10,000 OpenVLA-OFT and 6,000 $\pi_{0.5}$ episodes at $B=64$, and LIBERO-Plus adds 5,600 episodes at each of $B=16$, $32$, and $64$ across the five configurations in Table~\mainref{tab:plus} (1,120 paired conditions per configuration), totaling 88,800 closed-loop episodes before counting the 200-condition diagnostic and ablation panels. Each success rate counts one episode per configuration, task, and initial state, including failures. Online MOSAIC-VLA has no tuned hyperparameters: development panels and offline budget selection use initial states 0--4 (200 conditions), while held-out budget evaluation uses disjoint initial states 5--24 (800 conditions).

We report three timing quantities: request latency (one end-to-end policy call including token compression, excluding one-time compilation), model time per episode (the sum of policy calls within an episode), and episode wall time (including simulator stepping). Sequence replay measures prepared inference on fixed recorded trajectories so all compared configurations process identical inputs, while closed-loop timing records requests over the states visited during rollout. All compressed configurations keep Native's action-execution horizon.

Request-level speedups carry over to full episodes (Table~\ref{tab:episode_cost}). Without the count bias, failed rollouts run until the step limit, so Uniform issues 7.2\% and 9.4\% more requests per episode than Native in OpenVLA-OFT and $\pi_{0.5}$ (on the 1,804 OpenVLA-OFT conditions solved by both Uniform and Uniform\corr{}, request counts differ by only 1.1\%). Restoring the count bias keeps mean requests within 2.8\% of Native and reduces estimated episode model time (mean requests times sequence-replay latency) by 45.6\% and 39.0\%. In $\pi_{0.5}$, adding the bias lowers estimated episode model time from 1.67\,s to 1.58\,s because fewer steps outweigh the negligible mask overhead.

\begin{table}[htbp]
\centering\small
\caption{Requests and estimated model time per episode on the full benchmark at $B=64$ (execution seed 7, 2,000 episodes per configuration). Model time multiplies mean requests per episode by sequence-replay latency (for OpenVLA-OFT, Table~\ref{tab:e20} covers Native and the corrected methods).}
\label{tab:episode_cost}
\setlength{\tabcolsep}{3pt}

\begin{tabular}{@{}llrrrr@{}}
\toprule
Model & Method & Success (\%) & Requests & vs.\ Native & Model time (s) \\
\midrule
OpenVLA-OFT & Native & 97.25 & 19.75 & -- & 1.70 \\
 & Uniform & 92.05 & 21.18 & +7.2\% & -- \\
 & Uniform\corr{} & 96.65 & 19.68 & $-$0.3\% & 0.92 \\
 & Tree & 94.50 & 20.30 & +2.8\% & -- \\
 & Tree\corr{} & 96.45 & 19.79 & +0.2\% & 0.95 \\
\midrule
$\pi_{0.5}$ & Native & 96.85 & 32.40 & -- & 2.59 \\
 & Uniform & 94.80 & 35.46 & +9.4\% & 1.67 \\
 & Uniform\corr{} & 96.50 & 33.31 & +2.8\% & 1.58 \\
 & Tree & 95.25 & 34.62 & +6.8\% & 1.64 \\
 & Tree\corr{} & 96.80 & 32.75 & +1.1\% & 1.55 \\
\bottomrule
\end{tabular}

\end{table}

On 20 saved LIBERO-Spatial inputs (320 timed calls), replaying Uniform, zero-mask Uniform, and Uniform\corr{} averages 46.69, 47.01, and 47.00\,ms, and the corresponding Tree runs average 47.87, 48.43, and 48.49\,ms, confirming that constructing and applying the additive bias adds negligible latency (expanded $\pi_{0.5}$ readout diagnostics follow in Sec.~\ref{app:readout_metrics}).
\begin{table}[htbp]
\centering\small
\caption{$\pi_{0.5}$ partition rules at $B=64$ on the full benchmark (2,000 episodes per cell), without (``None'') and with the count bias, alongside observed closed-loop mean request times.}
\label{tab:allocations}
\begin{tabular}{lrrrr}
\toprule
& \multicolumn{2}{c}{Success (\%)} & \multicolumn{2}{c}{ms/request} \\ Rule & None & \corr{} & None & \corr{} \\
\midrule
Uniform & 94.80 & 96.50 & 40.14 & 40.04 \\
Tree & 95.25 & 96.80 & 40.17 & 40.15 \\
Greedy & 96.00 & 96.40 & 40.32 & 40.25 \\
DP & 95.30 & 96.10 & 40.25 & 40.21 \\
\bottomrule
\end{tabular}

\end{table}

\subsection{OFT Partition and Overhead Diagnostics}
\label{app:diagnostics}
On 4,000 first-observation camera grids in OpenVLA-OFT (2,000 conditions per camera, Table~\ref{tab:geometry}), Tree reallocates an average of 10.90 and 7.14 refinement splits relative to the equal-depth Uniform grid, changing 30.49 and 20.42 region representatives in the main and wrist views and lowering within-region distortion on every view. Component timing inside the end-to-end request call shows that Uniform token packing takes 0.05--0.06\,ms per request, Tree partitioning and packing take 1.25--1.35\,ms, and count-bias mask construction takes about 0.12\,ms (Table~\ref{tab:overhead}).

\begin{table}[htbp]
\centering\small
\caption{OpenVLA-OFT first-observation Tree partition statistics at $B=64$ (2,000 conditions per view).}
\label{tab:geometry}
\begin{tabular}{lrr}\toprule
Statistic & Camera 0 & Camera 1\\\midrule
Standard regions (\%) & 52.36 & 68.09\\
Changed representatives & 30.49 & 20.42\\
Extra refinement tokens & 10.90 & 7.14\\
\bottomrule\end{tabular}

\end{table}

\begin{table}[htbp]
\centering\small
\caption{OpenVLA-OFT per-request compression overhead (ms), included in total request latency.}
\label{tab:overhead}
\begin{tabular}{lrr}\toprule
Method & Pack & Correction mask \\\midrule
Native & 0.000 & 0.000\\
\midrule
Uniform & 0.061 & 0.000\\
Tree & 1.245 & 0.000\\
Uniform\corr{} & 0.053 & 0.116\\
Tree\corr{} & 1.349 & 0.120\\
\bottomrule\end{tabular}

\end{table}

\subsection{Sequence-Replay Request Timing}
\label{app:timing}
Table~\ref{tab:e20} reports sequence-replay latency for OpenVLA-OFT and the public pruners, and Table~\ref{tab:pi05_sequence} reports the corresponding measurements across all 17 $\pi_{0.5}$ configurations. At 64 total tokens, Uniform\corr{} and Tree\corr{} serve OpenVLA-OFT requests $1.88\times$ and $1.85\times$ faster than Native, while in $\pi_{0.5}$ adding the count bias changes mean request time by at most 0.18\,ms at any budget. Replaying uncorrected OpenVLA-OFT Uniform at $B=64$ on the same 40 recorded sequences in an additional RTX~5090 timing run averages 48.07\,ms per request ($1.81\times$ relative to its same-run 87.13\,ms Native reference, and $1.79\times$ relative to the 85.99\,ms Native mean of Table~\ref{tab:e20} used across Table~\mainref{tab:main}a).
\begin{table}[htbp]
\centering\small
\caption{Sequence-replay request latency for OpenVLA-OFT and public pruners on one RTX 5090 (1,530 requests per configuration: 40 trajectories recorded from 128-token SAFE-Pruner rollouts, replayed twice with state progression after warmup, first requests included). Each method runs in its upstream implementation in separate load blocks. Speedup is Native mean over configuration mean, TFLOPs count covered dense-equivalent matrix, convolution, and attention operations ($10^{12}$ per request), and Aux.\ counts auxiliary forwards (included in SAFE-Pruner's p95 on sequence-initial requests).}
\label{tab:e20}
\adjustbox{max width=\linewidth}{
\begin{tabular}{@{}lrrrrrr@{}}
\toprule
Method & Total tokens & Mean (ms) & p95 (ms) & Speedup & TFLOPs & Aux. \\
\midrule
Native & 512 & 85.99 & 87.85 & 1.00\(\times\) & 8.94 & 0 \\
U+L B32 & 64 & 45.67 & 47.63 & 1.88\(\times\) & 2.85 & 0 \\
T+L B32 & 64 & 46.44 & 48.50 & 1.85\(\times\) & 2.85 & 0 \\
U+L B64 & 128 & 46.93 & 48.87 & 1.83\(\times\) & 3.70 & 0 \\
T+L B64 & 128 & 48.15 & 50.20 & 1.79\(\times\) & 3.70 & 0 \\
ADP & 64 & 45.98 & 47.30 & 1.87\(\times\) & 2.86 & 0 \\
ADP & 128 & 46.82 & 48.54 & 1.84\(\times\) & 3.72 & 0 \\
SAFE & 64 & 57.79 & 140.55 & 1.49\(\times\) & 4.12 & 122 \\
SAFE & 128 & 59.17 & 142.05 & 1.45\(\times\) & 4.90 & 122 \\
VLA-Pruner & 64 & 54.08 & 55.25 & 1.59\(\times\) & 3.59 & 0 \\
VLA-Pruner & 128 & 56.59 & 58.15 & 1.52\(\times\) & 4.35 & 0 \\
\bottomrule
\end{tabular}
}
\end{table}
\begin{table}[htbp]
\centering\small
\caption{Sequence-replay request latency of $\pi_{0.5}$ on one RTX 5080 (ms, mean / p95) over 2,518 requests per configuration (40 Native trajectories totaling 1,259 requests, replayed twice after warmup, first requests included). All 17 configurations share one compiled runtime and time full synchronized inference calls including action host transfer.}
\label{tab:pi05_sequence}
\adjustbox{max width=\linewidth}{\begin{tabular}{lrrrr}\toprule
$B$ & Uniform & Uniform\corr{} & Tree & Tree\corr{}\\\midrule
Native & \multicolumn{4}{c}{79.97 / 81.35}\\
16 & 42.33 / 43.98 & 42.43 / 43.94 & 42.42 / 43.77 & 42.52 / 44.03\\
32 & 42.70 / 44.15 & 42.67 / 43.98 & 42.69 / 44.13 & 42.75 / 44.09\\
64 & 47.23 / 48.64 & 47.41 / 48.83 & 47.29 / 48.58 & 47.25 / 48.69\\
128 & 53.28 / 54.61 & 53.39 / 54.71 & 53.38 / 54.64 & 53.44 / 54.82\\
\bottomrule
\end{tabular}
}
\end{table}

\subsection{Comparison Families and Baseline Scope}
\label{app:pi05ports}
\paragraph{Prespecified Comparison Families.}
On the 200-condition development panels, confidence levels follow Bonferroni adjustment within each study's comparison family: the correction-path study at $B=64$ tests one primary contrast (95\%) and four secondary contrasts (98.75\%, while the $B=16$ route contrasts report nominal 95\% task-cluster bootstrap intervals), the four-budget OpenVLA-OFT partition study tests four primary contrasts (98.75\%), the three-budget OpenVLA-OFT similarity-merging study tests three primary contrasts (98.33\%) and six secondary contrasts (99.17\%), the $B=32$ $\pi_{0.5}$ merging study tests one primary contrast (95\%) and three secondary contrasts (98.33\%), and the $\pi_{0.5}$ interaction contrasts form a separate family.

\paragraph{Public Implementation Scope.}
Table~\mainref{tab:main}b evaluates the author-released OpenVLA-OFT implementations of ADP, SAFE-Pruner, and VLA-Pruner under our execution-seed protocol at fixed per-query budgets of 64 and 128 total visual tokens, keeping each method's token-selection rule while disabling full-token fallback so every request runs at the target budget. Because the public repositories of all three methods release OpenVLA-OFT code only (SAFE-Pruner's paper also reports $\pi_{0.5}$ results~\cite{safe}, without a public $\pi_{0.5}$ release), pruner comparisons use OpenVLA-OFT, where each repository's unpruned reference matches Native within 0.8 percentage points.

\section{Mechanism, Ablations, and Additional Results}
\label{supp:mechanism}

\subsection{Saved-Input Readout Measurements}
\label{app:readout_metrics}
We measure visual attention share error against same-input Native across all expert cells. Let $p_{\ell hf}$ denote the mean visual attention probability over the first five action queries at expert layer $\ell\in\{1,\dots,18\}$, head $h\in\{1,\dots,8\}$, and flow step $f\in\{1,\dots,10\}$, with $\langle\cdot\rangle_{\ell hf}$ averaging across all $18\times8\times10$ cells:
\[
D_{\mathrm{cell}}=\bigl\langle |p_{\ell hf}-p^{\mathrm{Native}}_{\ell hf}|\bigr\rangle_{\ell hf}.
\]
For each cell $(\ell,h,f)$, let $\bar m_{\ell hf}=\frac{1}{5}\sum_{q=1}^{5}\sum_{j\in\mathcal V} a_{\ell hf,qj}v_{\ell hf,j}$ be the query-averaged visual message vector: visual-message RMSE averages the RMSE between compressed and Native $\bar m_{\ell hf}$ across cells, and translation-command RMSE compares the first five predicted $(\Delta x,\Delta y,\Delta z)$ commands against same-input Native. Across all saved-input diagnostics, attention hooks leave predicted actions bitwise unchanged, and repeated instrumented passes agree exactly.

\subsubsection{First-Observation Diagnostic: 200 Observations}
On 200 saved first observations from recorded Native episodes (all 40 tasks, initial states 0--4), Table~\ref{tab:attention200} reports $D_{\mathrm{cell}}$, visual-message RMSE, and translation-command RMSE for Native and six $B=64$ compressed configurations in $\pi_{0.5}$. Correcting the action expert reduces $D_{\mathrm{cell}}$ from 0.1450 to 0.0434 under Uniform, while correcting both readout paths lowers message and action RMSE further, and Tree\corr{} achieves the lowest errors among all six compressed configurations across all three metrics (0.0325, 0.1136, and 0.0343). On the same 200 observations at $B=16$, Table~\ref{tab:counterfactual} compares translation-command RMSE in both $\pi_{0.5}$ and OpenVLA-OFT when diluting readout on all 256 tokens ($\gamma=0.25$) versus restoring Native share ($\gamma=1$) on 16 Uniform tokens per camera.

\begin{table}[htbp]
\centering\small
\caption{Saved-input readout diagnostic in $\pi_{0.5}$ at $B=64$ across 200 first observations (40 tasks, initial states 0--4) relative to same-input Native. Prefix, expert, and both indicate where $+\log|G|$ is applied.}
\label{tab:attention200}
\begin{tabular}{lrrr}
\toprule
Configuration & Share error & Message RMSE & Translation RMSE \\
\midrule
Native & 0.0000 & 0.0000 & 0.0000 \\
Uniform & 0.1450 & 0.1583 & 0.0555 \\
Uniform, both & 0.0456 & 0.1313 & 0.0428 \\
Uniform, prefix & 0.1514 & 0.1549 & 0.0491 \\
Uniform, expert & 0.0434 & 0.1434 & 0.0493 \\
Tree & 0.1206 & 0.1460 & 0.0449 \\
Tree, \corr{} & 0.0325 & 0.1136 & 0.0343 \\
\bottomrule
\end{tabular}

\end{table}

\begin{table}[htbp]
\centering\small
\caption{Translation-command RMSE against same-input Native at $B=16$ on 200 saved observations ($\pi_{0.5}$ / OpenVLA-OFT), crossing full (256) versus merged (16 Uniform) tokens with restored ($\gamma=1$) versus diluted ($\gamma=0.25$ on Native, or uncorrected merging) readout.}
\label{tab:counterfactual}
\setlength{\tabcolsep}{4pt}
\begin{tabular}{lcc}\toprule
Visual tokens & Readout restored & Readout diluted\\\midrule
All 256 per camera & 0 (Native) & 0.073 / 0.109\\
16 merged per camera & 0.076 / 0.109 & 0.150 / 0.197\\
\bottomrule
\end{tabular}

\end{table}

\subsubsection{Intervention on Action-Expert Visual Share}
\label{app:native_share}
To test whether prefix correction still helps once the action expert's visual share matches Native, a second $B=64$ experiment fixes the Uniform partition on the same 200 observations and steers every expert layer, head, flow step, and action query ($1,\dots,10$) to its same-input Native visual share $\rho$. Given a rounded attention row $a$ with $A_V=\sum_{j\in\mathcal V}a_j$ and $A_O=\sum_{j\notin\mathcal V}a_j$ (when $A_V,A_O>0$), we rescale in FP32 and cast back to BF16:
\[
\widetilde a_j=
\begin{cases}
\rho\,a_j/A_V, & j\in\mathcal V,\\
(1-\rho)\,a_j/A_O, & j\notin\mathcal V,
\end{cases}
\]
matching the target normalized share $A_V/(A_V+A_O)$ to a maximum observed error of 0.00141 while preserving masked zeros and within-group relative weights up to BF16 rounding. Crossing Native-share steering with prefix correction off or on, alongside self-share rescaling controls (which set $\rho$ to each row's own current normalized share, inducing at most 0.000354 action RMSE from BF16 recasting) and the unmodified paths, yields the eight configurations in Table~\ref{tab:mass200}.

Using 10,000 task-cluster bootstrap draws within suites (seed 20261027), Table~\ref{tab:mass200_contrasts} evaluates the prespecified primary contrast (prefix correction on minus off under Native-share steering, 95\% interval) and two secondary Native-share versus self-share contrasts (97.5\% intervals). All three intervals lie wholly below zero, with prefix correction reducing command error in every suite even when the expert's scalar visual share is already matched to Native, reflecting improved multimodal prefix representations.

\begin{table}[htbp]
\centering\small
\caption{Action-expert Native-share intervention on 200 first $\pi_{0.5}$ observations with Uniform at $B=64$ (errors relative to same-input Native, where Share error is raw $D_{\mathrm{cell}}$ and the intervention targets normalized share $A_V/(A_V+A_O)$).}
\label{tab:mass200}
\begin{tabular}{lrrr}
\toprule
Configuration & Share error & Message RMSE & Translation RMSE \\
\midrule
Native & 0.0000 & 0.0000 & 0.0000 \\
Uniform & 0.1450 & 0.1583 & 0.0555 \\
Uniform, both & 0.0456 & 0.1313 & 0.0428 \\
Uniform, prefix & 0.1514 & 0.1549 & 0.0491 \\
No prefix, self share & 0.1450 & 0.1583 & 0.0555 \\
Prefix, self share & 0.1514 & 0.1549 & 0.0491 \\
No prefix, Native share & 0.0001 & 0.1397 & 0.0471 \\
Prefix, Native share & 0.0001 & 0.1312 & 0.0410 \\
\bottomrule
\end{tabular}

\end{table}

\begin{table}[htbp]
\centering\small
\caption{Prespecified translation-command RMSE contrasts for the Native-share intervention (Table~\ref{tab:mass200}). Row 1 is the primary contrast (95\% interval), and rows 2--3 form the secondary family (97.5\% intervals), where negative values denote lower error.}
\label{tab:mass200_contrasts}
\begin{tabular}{lrr}
\toprule
Contrast & Difference & Confidence interval \\
\midrule
Prefix effect at Native share & $-0.0061$ & $[-0.0083, -0.0039]$ (95\%) \\
Native vs. self share, no prefix & $-0.0084$ & $[-0.0119, -0.0050]$ (97.5\%) \\
Native vs. self share, prefix & $-0.0081$ & $[-0.0119, -0.0044]$ (97.5\%) \\
\bottomrule
\end{tabular}

\end{table}

\subsection{Controlled-Share Cells}
\label{app:prediction}
Table~\ref{tab:gamma_cells} lists every controlled-share cell, including OpenVLA-OFT at $B=24$, which Fig.~\mainref{fig:law}c leaves out for legibility.

\begin{table}[htbp]
\centering\footnotesize
\caption{\textbf{Success (\%) with the action path's visual share set to $\gamma$ times Native.} Fixed Uniform\corr{} tokens, 200 conditions per cell. At $B=24$ and 48, Uniform uses balanced $6\times4$ and $8\times6$ grids of rectangles with their true member counts. At $\gamma=0$ the action path reads no visual token directly.}
\label{tab:gamma_cells}
\setlength{\tabcolsep}{3pt}
\begin{tabular}{llrrrrrr}\toprule
&& \multicolumn{6}{c}{$\gamma$}\\\cmidrule(lr){3-8}
Model & $B$ & 0 & 0.25 & 0.5 & 0.75 & 0.9 & 1\\\midrule
OpenVLA-OFT & 16 & -- & 5.0 & 36.0 & 56.5 & 67.0 & 62.0\\
OpenVLA-OFT & 24 & -- & 16.0 & 55.5 & 77.5 & 84.0 & 85.0\\
OpenVLA-OFT & 32 & -- & 26.5 & 72.5 & 88.0 & 90.5 & 92.5\\
OpenVLA-OFT & 48 & -- & 29.5 & 82.0 & 95.5 & 96.5 & 97.0\\
OpenVLA-OFT & 64 & 0.0 & -- & -- & -- & -- & 96.5\\
OpenVLA-OFT & 128 & 0.0 & -- & -- & -- & -- & 95.0\\
$\pi_{0.5}$ & 32 & -- & 19.5 & 74.0 & 87.0 & 88.5 & 83.5\\
$\pi_{0.5}$ & 64 & -- & 38.0 & 91.5 & 94.5 & 96.0 & 95.0\\
\bottomrule
\end{tabular}

\end{table}

\subsection{Partitions, Corrected Readout, and Action Error}
\label{app:partition_readout}
We test prediction (iii) on the 200 saved $\pi_{0.5}$ inputs (40 tasks, initial states 0--4) with Uniform, Tree, Greedy, and fixed-tree DP at $B=16$, 32, and 64, all with the bias on both paths and with each input's observation and flow noise held fixed (Table~\ref{tab:partition_readout} and Fig.~\ref{fig:partition_readout}).
The shift is the median visual log-odds change against Native over the action expert's 18 layers, 10 flow steps, eight heads, and first five action queries.
Action error is the mean squared difference of the first five translation commands from Native's.
At $B=64$, all three adaptive partitions leave a smaller corrected shift than Uniform, with fixed-tree DP achieving the smallest shift: Tree, Greedy, and DP shrink the shift magnitude by 36\%, 25\%, and 41\% and lower action error by 32\%, 38\%, and 52\%. At $B=16$ and $B=32$, DP continues to reduce action error by 18\% and 16\% relative to Uniform\corr{}. Median nonvisual log-mass changes remain within 0.02 nats across all configurations, confirming that the corrected log-odds shift is almost purely visual.

\begin{table}[htbp]
\centering\footnotesize
\caption{\textbf{Partitions in $\pi_{0.5}$ with the bias on both paths, 200 saved inputs.} $\Sigma E$ is distortion relative to one region per camera, Share the action expert's mean visual attention (Native 31.7\%), Shift the corrected visual log-odds shift against Native, and Err.\ the translation error against Native. $\Delta$ columns give the change relative to Uniform at the same budget, with 95\% task-cluster intervals.}
\label{tab:partition_readout}
\setlength{\tabcolsep}{5pt}
\begin{tabular}{rlrrrrrr}\toprule
$B$ & Rule & $\Sigma E$ (rel.) & Share (\%) & Shift & $\Delta$ vs.\ U (\%) & Err.\ ($10^{-3}$) & $\Delta$ vs.\ U (\%)\\\midrule
16 & Uniform & 0.626 & 28.2 & $-0.334$ & -- & 8.70 & --\\
 & Tree & 0.619 & 27.9 & $-0.337$ & $+1$ {\scriptsize$[-1,+3]$} & 7.13 & $-18$ {\scriptsize$[-29,-7]$}\\
 & Greedy & 0.617 & 27.7 & $-0.350$ & $+5$ {\scriptsize$[+2,+8]$} & 7.48 & $-14$ {\scriptsize$[-27,0]$}\\
 & DP & 0.614 & 27.6 & $-0.358$ & $+7$ {\scriptsize$[+5,+10]$} & 7.15 & $-18$ {\scriptsize$[-33,-3]$}\\
\midrule
32 & Uniform & 0.516 & 28.4 & $-0.284$ & -- & 4.13 & --\\
 & Tree & 0.495 & 27.7 & $-0.336$ & $+18$ {\scriptsize$[+16,+21]$} & 4.12 & $-0.3$ {\scriptsize$[-12,+10]$}\\
 & Greedy & 0.484 & 27.8 & $-0.309$ & $+9$ {\scriptsize$[+6,+12]$} & 3.69 & $-11$ {\scriptsize$[-30,+11]$}\\
 & DP & 0.478 & 27.9 & $-0.303$ & $+7$ {\scriptsize$[+3,+10]$} & 3.47 & $-16$ {\scriptsize$[-29,-2]$}\\
\midrule
64 & Uniform & 0.370 & 28.2 & $-0.289$ & -- & 2.43 & --\\
 & Tree & 0.348 & 29.4 & $-0.186$ & $-36$ {\scriptsize$[-38,-33]$} & 1.64 & $-32$ {\scriptsize$[-44,-19]$}\\
 & Greedy & 0.341 & 29.0 & $-0.216$ & $-25$ {\scriptsize$[-28,-23]$} & 1.51 & $-38$ {\scriptsize$[-51,-22]$}\\
 & DP & 0.328 & 29.5 & $-0.171$ & $-41$ {\scriptsize$[-43,-38]$} & 1.16 & $-52$ {\scriptsize$[-60,-42]$}\\
\bottomrule
\end{tabular}

\end{table}

\begin{figure}[htbp]
\centering
\captionsetup{skip=4pt}
\includegraphics[width=0.90\textwidth]{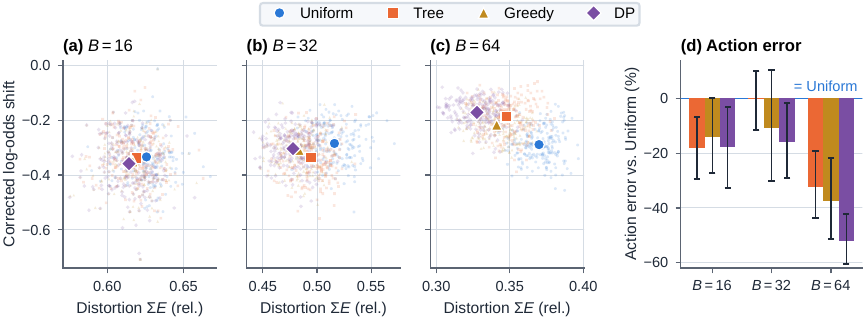}
\caption{\textbf{Distortion, corrected readout, and action error in $\pi_{0.5}$.} (a--c) Relative distortion against the corrected visual log-odds shift for four partitions with the bias, per saved input (small markers) and on average (large markers). (d) Translation error against Native relative to Uniform at the same budget, with 95\% task-cluster intervals.}
\label{fig:partition_readout}
\end{figure}

\subsection{Where the Action Expert Reads}
\label{app:heatmaps}
Fig.~\ref{fig:heatmaps} places this attention on image patches for two saved inputs fixed before measurement, the first request of task 0 from initial state 0 in LIBERO-Spatial and LIBERO-Long.
Each patch shows the action expert's attention probability averaged over the first five action queries, eight heads, 18 layers, and ten flow steps, and a merged token's probability is spread evenly over its member patches for display.
In Native, 77\% and 79\% of the visual attention falls on the wrist camera and about half on the bottom quarter of the wrist image, where the gripper fingers appear.
Diluting Native to $\gamma=0.242$ keeps this pattern at 0.24 of its weight.
Uniform merging at $B=16$ leaves 0.18 and 0.29 of Native's visual share, and the bias restores 0.86 and 0.97, again concentrated on the gripper region.

\begin{figure}[htbp]
\centering
\captionsetup{skip=4pt}
\includegraphics[width=0.84\textwidth]{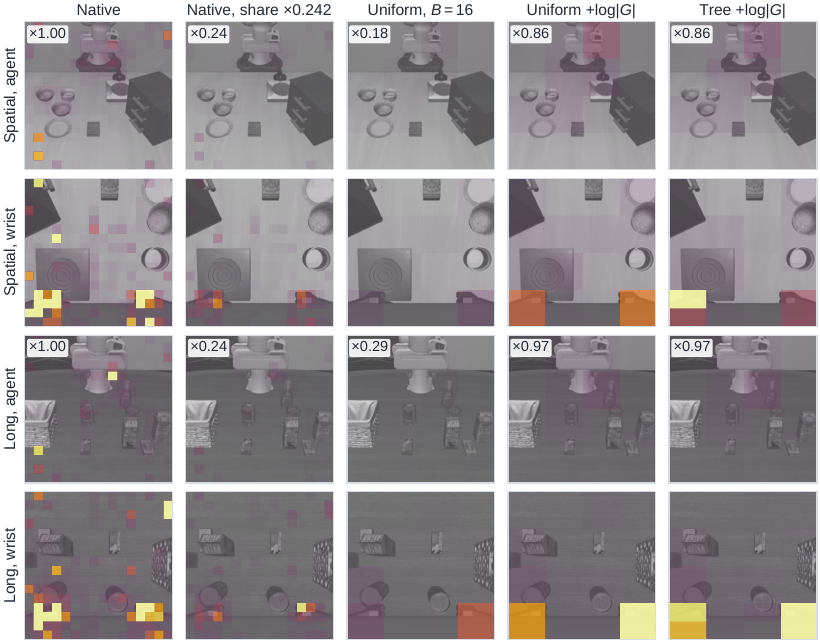}
\caption{\textbf{Where $\pi_{0.5}$'s action expert reads.} Attention probability per image patch on two saved inputs (rows: suite and camera), on one color scale per input, for Native, Native diluted to $\gamma=0.242$, and Uniform merging at $B=16$ without and with the bias, and Tree\corr{}. Labels give each setting's total visual share relative to Native on that input. A merged token's probability is spread evenly over its member patches.}
\label{fig:heatmaps}
\end{figure}

\subsection{Where the Lost Visual Share Goes}
\label{app:keytypes}
On 200 saved $\pi_{0.5}$ observations at $B=64$, Table~\ref{tab:keytypes} splits the action path's attention over four key types: visual keys, the first language key, identified by its fixed position before measurement, the other language keys, and the action keys. Shares average over queries, heads, layers, and flow steps.
In Native the first language key alone holds 34.9\% of this attention, more than all visual keys together (31.7\%), as expected of an attention sink~\cite{streamingllm}.
Uncorrected Uniform lowers the visual share to 17.2\%, while the first language key, the remaining language keys, and the action keys gain 4.53, 4.64, and 5.31 percentage points, spreading the lost visual share across all nonvisual token groups. Adding the count bias restores the visual share to 28.2\%.

\begin{table}[htbp]
\centering\small
\caption{Action-path attention (\%) of $\pi_{0.5}$ by key type on 200 saved observations at $B=64$, with the change under uncorrected Uniform and its 95\% task-cluster interval.}
\label{tab:keytypes}
\adjustbox{max width=\linewidth}{
\begin{tabular}{@{}lrrrr@{}}
\toprule
Key type & Native & Uniform & Uniform\corr{} & Uniform $-$ Native \\
\midrule
Visual keys & 31.7 & 17.2 & 28.2 & $-$14.47 [$-$14.62, $-$14.33] \\
First language key & 34.9 & 39.4 & 35.8 & 4.53 [4.38, 4.67] \\
Other language keys & 8.8 & 13.5 & 10.3 & 4.64 [4.50, 4.79] \\
Action keys & 24.6 & 29.9 & 25.8 & 5.31 [5.22, 5.39] \\
\bottomrule
\end{tabular}
}
\end{table}

\subsection{Correction-Path Evaluation Protocol}
\label{app:readout_closed_loop}
To isolate the two readout paths of $\pi_{0.5}$ in closed loop, we evaluate four Uniform merging routes (bias on neither path, prefix only, action expert only, or both) across the 200 paired development conditions (all 40 tasks, initial states 0--4, 800 episodes per panel) on one RTX~5080 with fresh per-episode simulator processes, identical initial states and flow-noise schedules, ten predicted actions, five executed actions per request, ten flow steps, and 64 language slots, computing all paired contrasts within each study's own rollouts. Paired differences average equally across tasks and suites using 10,000 task-cluster bootstrap draws within suites (seed 20261025). At $B=64$, the four routes succeed in 188, 188, 192, and 193 of 200 trials (94.0\%, 94.0\%, 96.0\%, and 96.5\%, Table~\ref{tab:readout_closed_loop_contrasts}): both-minus-expert-only is the primary contrast (95\% interval), the remaining four comparisons form the secondary family (98.75\% intervals), and among the 13 conditions where uncorrected merging and dual-path correction disagree, dual-path correction wins 9.

\begin{table}[htbp]
\centering\small
\caption{\textbf{Prespecified correction-path contrasts in $\pi_{0.5}$ at $B=64$} (percentage points, 200 paired conditions). Row 1 is the primary contrast (95\%) and rows 2--5 form the secondary family (98.75\%), where Interaction is both $-$ prefix-only $-$ expert-only $+$ none.}
\label{tab:readout_closed_loop_contrasts}
\begin{tabular}{lrrr}
\toprule
Comparison & Difference (pp) & CI level & Interval (pp) \\
\midrule
Both $-$ expert-only & $+0.5$ & 95\% & $[-2.5,4.5]$ \\
Prefix-only $-$ none & $+0.0$ & 98.75\% & $[-5.5,4.5]$ \\
Expert-only $-$ none & $+2.0$ & 98.75\% & $[-3.5,7.5]$ \\
Both $-$ prefix-only & $+2.5$ & 98.75\% & $[-1.0,6.5]$ \\
Interaction & $+0.5$ & 98.75\% & $[-5.0,6.5]$ \\
\bottomrule
\end{tabular}

\end{table}

\paragraph{Closed-Loop Routes at $B=16$.}
Repeating the four routes at $B=16$ on the same 200 conditions yields 12.0\%, 32.5\%, 38.0\%, and 67.5\% success with the bias applied nowhere, on the prefix only, on the expert only, and on both paths (Table~\ref{tab:reading_routes}). Applying proportional attention only on the language prefix, analogous to single-backbone VLM merging~\cite{pact}, leaves the action expert diluted under strong compression (32.5\% vs.\ 67.5\% with both paths), whereas Table~\ref{tab:mass200_contrasts} confirms an independent prefix benefit under matched expert share and Table~\ref{tab:readout_closed_loop_contrasts} shows the closed-loop gap narrowing at $B=64$.

\begin{table}[htbp]
\centering\small
\caption{\textbf{Closed-loop readout routes in $\pi_{0.5}$ at $B=16$} (paired differences in percentage points across 200 conditions, 95\% task-cluster bootstrap intervals from 10,000 draws).}
\label{tab:reading_routes}
\begin{tabular}{lrr}\toprule
Contrast & Difference (pp) & 95\% interval\\\midrule
Prefix $-$ none & $+20.5$ & $[13.0,28.0]$\\
Expert $-$ none & $+26.0$ & $[17.0,35.0]$\\
Both $-$ none & $+55.5$ & $[45.0,65.0]$\\
Both $-$ prefix & $+35.0$ & $[24.5,45.0]$\\
Both $-$ expert & $+29.5$ & $[19.0,40.0]$\\
Interaction & $+9.0$ & $[-5.0,22.5]$\\
\bottomrule
\end{tabular}

\end{table}

\subsection{Where the Bias Is Applied}
\label{app:placement}
This ablation merges inside $\pi_{0.5}$'s 27-layer SigLIP encoder, before layer 26, with the Tree partition at $B=16$, so the last two encoder layers already see merged tokens. Four arms share the merge position, the groups, and $\alpha=1$, and form a $2\times2$ design over where $\log s_G$ enters: nowhere, in the remaining encoder layers only, on the prefix and action-expert readout paths only, or in both. Each arm runs the 200 development conditions (40 tasks, initial states 0--4) with execution seed 7 (Table~\ref{tab:placement}). The three contrasts against no correction and the interaction form one comparison family with Bonferroni-adjusted task-cluster intervals (5,000 draws, seed 20261101).

Correcting only inside the encoder leaves success at the uncorrected level (31.5\% against 30.5\%, $+1.0$ points, $[-5.5,7.5]$). Correcting the readout paths gives 74.5\%, $+44.0$ points ($[31.0,57.0]$), and adding the encoder correction on top does not add to it (73.0\%, $+42.5$ points, $[29.5,55.0]$, interaction $-2.5$ points, $[-11.5,6.0]$). Size-weighted attention inside the encoder, as in encoder-side merging~\cite{tome,dymu}, therefore does not substitute for restoring the count where the policy reads.

\begin{table}[htbp]
\centering\small
\caption{Correction placement in $\pi_{0.5}$ at $B=16$ with merging inside the vision encoder ($2\times2$ over encoder and readout correction), 200 conditions per configuration. Success (\%) with 95\% task-cluster intervals.}
\label{tab:placement}
\setlength{\tabcolsep}{3pt}
\adjustbox{max width=\linewidth}{
\begin{tabular}{@{}lccr@{}}
\toprule
Configuration & Bias in encoder & Bias on readout & Success (\%) \\
\midrule
No correction & no & no & 30.5 [22.0, 39.5] \\
Encoder only & yes & no & 31.5 [23.0, 40.5] \\
Readout only & no & yes & 74.5 [65.0, 83.0] \\
Both & yes & yes & 73.0 [63.0, 82.0] \\
\bottomrule
\end{tabular}
}
\end{table}

\subsection{Second Execution Seed on the Full Benchmark}
\label{app:seed13}
The full $B=64$ benchmark is repeated with execution seed 13 on the same 2,000 conditions per configuration (40 tasks with 50 released initial states each), whereas the main results use seed 7. OpenVLA-OFT runs the simulator with environment seeds 0 and 13 for the two executions and $\pi_{0.5}$ with 7 and 13, and checkpoints, preprocessing, merging, and the count bias are unchanged. Because the two executions share the same 2,000 conditions, Table~\ref{tab:seed_replication} reports each execution alongside their mean.

Seed 13 reproduces the seed-7 pattern. Uniform\corr{} is 0.80 points below Native in OpenVLA-OFT and 0.25 points above it in $\pi_{0.5}$, against 0.60 and 0.35 points below with seed 7. The bias adds 3.85 and 2.80 points to Uniform (4.60 and 1.70 with seed 7), and 1.65 points to Tree in OpenVLA-OFT (1.95). Per condition, Uniform\corr{} succeeds where Uniform fails in 119 and 94 conditions, and the reverse holds in 42 and 38. The gain is again largest on LIBERO-Long: 8.8 points in OpenVLA-OFT (82.6\% to 91.4\%) and 6.2 in $\pi_{0.5}$ (88.6\% to 94.8\%), against 10.6 and 3.6 with seed 7.

\begin{table}[htbp]
\centering\small
\caption{Full-benchmark success (\%) at $B=64$ under two execution seeds, 2,000 conditions per configuration and seed. Seed 7 is Table~\mainref{tab:main}a. In $\pi_{0.5}$, the second seed covers Native, Uniform, and Uniform\corr{}.}
\label{tab:seed_replication}
\setlength{\tabcolsep}{4.5pt}

\begin{tabular}{@{}llrrr@{}}
\toprule
Model & Method & Seed 7 & Seed 13 & Mean \\
\midrule
OpenVLA-OFT & Native & 97.25 & 96.90 & 97.08 \\
 & Uniform & 92.05 & 92.25 & 92.15 \\
 & Uniform\corr{} & 96.65 & 96.10 & 96.38 \\
 & Tree & 94.50 & 94.70 & 94.60 \\
 & Tree\corr{} & 96.45 & 96.35 & 96.40 \\
\midrule
$\pi_{0.5}$ & Native & 96.85 & 96.50 & 96.68 \\
 & Uniform & 94.80 & 93.95 & 94.38 \\
 & Uniform\corr{} & 96.50 & 96.75 & 96.63 \\
\bottomrule
\end{tabular}

\end{table}

\subsection{Full-Benchmark Success by Suite}
\label{app:bysuite}
Table~\ref{tab:success_full} breaks the full-benchmark results down by suite. With the count bias, both partitions stay within 1.8 points of Native in every suite of both models, and the bias gains most on LIBERO-Long, where it lifts Uniform from 82.8\% to 93.4\% in OpenVLA-OFT and from 89.4\% to 93.0\% in $\pi_{0.5}$.
\begin{table}[htbp]
\centering\small
\caption{Full-benchmark success (\%) at $B=64$ by LIBERO suite for every partition with and without the count bias. Each suite has 500 episodes per configuration and the average 2,000. Rows within a model share conditions. Long is LIBERO-10.}
\label{tab:success_full}
\setlength{\tabcolsep}{4pt}
\begin{tabular}{lrrrrr}\toprule
Configuration & Spatial & Object & Goal & Long & Avg.\\\midrule
\multicolumn{6}{l}{\textit{OpenVLA-OFT}}\\
Native & 97.6 & 97.6 & 98.6 & 95.2 & 97.25\\
Uniform & 95.4 & 97.8 & 92.2 & 82.8 & 92.05\\
Tree & 96.8 & 96.6 & 94.6 & 90.0 & 94.50\\
Uniform\corr{} & 97.2 & 99.0 & 97.0 & 93.4 & 96.65\\
Tree\corr{} & 97.4 & 97.6 & 96.8 & 94.0 & 96.45\\
\midrule
\multicolumn{6}{l}{\textit{$\pi_{0.5}$}}\\
Native & 99.0 & 98.4 & 98.0 & 92.0 & 96.85\\
Uniform & 97.4 & 97.0 & 95.4 & 89.4 & 94.80\\
Tree & 98.0 & 97.4 & 98.0 & 87.6 & 95.25\\
Uniform\corr{} & 98.2 & 97.0 & 97.8 & 93.0 & 96.50\\
Tree\corr{} & 98.4 & 97.4 & 98.6 & 92.8 & 96.80\\
\bottomrule
\end{tabular}

\end{table}

\subsection{Three-Budget Similarity Merging}
\label{app:oft_tome}
We evaluate within-camera bipartite similarity merging (ToMe) with and without the count bias across $B\in\{16,32,64\}$ in OpenVLA-OFT and $\pi_{0.5}$ (Table~\mainref{tab:transfer}). In OpenVLA-OFT, the six ToMe configurations run on the 200 shared development conditions (40 tasks, initial states 0--4, seed 7, eight executed actions per request) paired with the Uniform\corr{} and Tree\corr{} episodes of the four-budget study from identical initial simulation states, observations, and instructions (Table~\ref{tab:oft_tome_main}). Within each camera's projected grid, the adapter performs size-weighted bipartite matching, assigns each merged region its minimum member position, preserves camera order and nonvisual positions, and optionally adds $+\log|G|$ across all language layers.

\begin{table}[htbp]
\centering\small
\caption{Prespecified OpenVLA-OFT similarity-merging contrasts on 200 paired conditions (U/T/M denote Uniform/Tree/ToMe, +L adds $+\log|G|$, and the last column counts discordant successes). Three primary comparisons use 98.33\% intervals and six secondary comparisons use 99.17\% intervals.}
\label{tab:oft_tome_contrasts}
\begin{tabular}{@{}rlrrrr@{}}
\toprule
$B$ & Contrast & Confidence & $\Delta$ (pp) & Adjusted interval (pp) & Only first / second \\
\midrule
16 & M+L $-$ U+L & 98.33\% & $-33.5$ & $[-43.00,-24.00]$ & 2 / 69 \\
32 & M+L $-$ U+L & 98.33\% & $-11.5$ & $[-18.50,-4.50]$ & 7 / 30 \\
64 & M+L $-$ U+L & 98.33\% & $+2.0$ & $[-1.00,5.50]$ & 5 / 1 \\
16 & T+L $-$ M+L & 99.17\% & $+52.5$ & $[41.50,62.00]$ & 105 / 0 \\
16 & M+L $-$ M & 99.17\% & $+10.0$ & $[3.00,18.00]$ & 20 / 0 \\
32 & T+L $-$ M+L & 99.17\% & $+19.5$ & $[9.50,29.50]$ & 42 / 3 \\
32 & M+L $-$ M & 99.17\% & $+32.5$ & $[20.50,44.50]$ & 67 / 2 \\
64 & T+L $-$ M+L & 99.17\% & $-1.5$ & $[-4.00,0.00]$ & 1 / 4 \\
64 & M+L $-$ M & 99.17\% & $+7.5$ & $[1.83,14.50]$ & 17 / 2 \\
\bottomrule
\end{tabular}

\end{table}

Resampling ten whole tasks per suite over 10,000 bootstrap draws (seed 20261027), Table~\ref{tab:oft_tome_contrasts} reports the three primary contrasts (ToMe\corr{} minus Uniform\corr{} at each budget, $98.33\%$ intervals) and six secondary contrasts comparing Tree\corr{} against ToMe\corr{} and ToMe\corr{} against uncorrected ToMe ($99.17\%$ intervals). Adding $+\log|G|$ improves ToMe at every budget, while Uniform\corr{} and Tree\corr{} lead ToMe\corr{} by wide margins at $B=16$ and $32$, and at $B=64$ ToMe\corr{} (198/200) and Uniform\corr{} (194/200) have an interval spanning zero.
\begin{table}[htbp]
\centering\small
\caption{OpenVLA-OFT similarity-merging success (\%) across 200 shared conditions (U/T denote Uniform/Tree, and +L adds $+\log|G|$).}
\label{tab:oft_tome_main}
\begin{tabular}{@{}rrrrr@{}}
\toprule
$B$ & U+L & T+L & ToMe & ToMe+L \\
\midrule
16 & 47.0 & 66.0 & 3.5 & 13.5 \\
32 & 87.0 & 95.0 & 43.0 & 75.5 \\
64 & 97.0 & 97.5 & 91.5 & 99.0 \\
\bottomrule
\end{tabular}

\end{table}

\paragraph{$\pi_{0.5}$ across $B\in\{16,32,64\}$.}
In $\pi_{0.5}$, all similarity-merging evaluations in Table~\mainref{tab:transfer} run on the same 200 shared conditions (40 tasks, initial states 0--4, seed 7, $H=10$, $n=5$, ten flow steps), with the $+\log|G|$ configurations applying the count bias along both readout paths. At $B=16$, ToMe succeeds in 6/200 trials (3.0\%) and ToMe\corr{} in 51/200 (25.5\%), a $+22.5$-point gain (95\% task-cluster interval $[15.5,30.0]$, 48 vs.\ 3 discordant conditions). At $B=32$, Table~\ref{tab:merge5080b32} gives the per-suite breakdown and Table~\ref{tab:merge5080b32_contrasts} reports the prespecified contrasts. At $B=64$, evaluating all five configurations together on one RTX~5090 (whereas full-benchmark rollouts and all $\pi_{0.5}$ latency measurements use the RTX~5080 setup of Sec.~\mainref{sec:eval}) yields 188, 192, 183, 184, and 191 successes out of 200 for Uniform, Uniform\corr{}, ToMe, ToMe\corr{}, and Tree\corr{} (94.0\%, 96.0\%, 91.5\%, 92.0\%, and 95.5\%).
\begin{table}[htbp]
\centering\small
\caption{$\pi_{0.5}$ similarity-merging panel at $B=32$ across the 200 shared conditions (50 per suite).}
\label{tab:merge5080b32}
\adjustbox{max width=\linewidth}{
\begin{tabular}{lrrrrr}\toprule
Configuration & Spatial & Object & Goal & Long & All\\\midrule
Uniform\corr{} & 40/50 & 49/50 & 44/50 & 39/50 & 172/200\\
Tree\corr{} & 46/50 & 48/50 & 45/50 & 40/50 & 179/200\\
ToMe & 29/50 & 30/50 & 42/50 & 15/50 & 116/200\\
ToMe\corr{} & 32/50 & 42/50 & 42/50 & 29/50 & 145/200\\
\bottomrule\end{tabular}
}
\end{table}
\begin{table}[htbp]
\centering\small
\caption{Prespecified $\pi_{0.5}$ contrasts at $B=32$ (percentage points, where U/T/M denote Uniform/Tree/ToMe and +L adds $+\log|G|$). Row 1 is primary (95\%), and rows 2--4 form the secondary family (98.33\%).}
\label{tab:merge5080b32_contrasts}
\begin{tabular}{lrl}\toprule
Contrast & Difference & Interval\\\midrule
T+L $-$ M+L & $+17.0$ & $[8.5,25.0]$\\
T+L $-$ U+L & $+3.5$ & $[-3.5,10.0]$\\
M+L $-$ U+L & $-13.5$ & $[-24.0,-1.5]$\\
M+L $-$ M & $+14.5$ & $[4.5,24.0]$\\
\bottomrule\end{tabular}

\end{table}

\subsection{Bias-Scale Sweep and Pruner-Merge Extension}
\label{app:alpha_adp}
On the 200 shared OpenVLA-OFT conditions, scaling the count bias by $\alpha$ ($\alpha\log s_G$) confirms that the theoretical unit weight ($\alpha=1$) needs no tuning: at $B=32$, Uniform with $\alpha=1$ reaches 87.0\% success versus 83.0\% at $\alpha=0.5$ and 81.5\% at $\alpha=1.5$, and at $B=64$ ($\alpha=1$ at 97.0\%) it lies within 0.5 percentage points of both $\alpha=0.5$ and $\alpha=1.5$. At $B=16$, increasing Uniform's scale to $\alpha=1.25$ or $\alpha=1.5$ remains below 50\%, whereas unscaled Tree\corr{} ($\alpha=1$) reaches 66.0\%, showing that rescaling a uniform grid's bias cannot substitute for adaptive partitioning. On the same 200 OpenVLA-OFT conditions, keeping ADP's token selection at 64 and 128 total tokens while merging each discarded patch into the most similar kept token of its camera and adding $+\log s_G$ raises ADP from 46.5\% and 64.5\% to 55.5\% and 70.5\%.

\subsection{Target Deployment}
\label{app:target}
Offline budget selection uses only initial states 0--4 (50 development episodes per suite) across Uniform\corr{} and Tree\corr{} at $B\in\{16,32,64,128\}$, allowing at most $\lfloor(\delta-1)/2\rfloor$ fewer successes than Native out of 50 (leaving a 1-point guard band for finite-sample estimation on 50 episodes). At $\delta=5$ percentage points, the rule selects Tree\corr{} $B=32$, Uniform\corr{} $B=32$, Tree\corr{} $B=32$, and Uniform\corr{} $B=64$ across Spatial, Object, Goal, and Long in OpenVLA-OFT, and Tree\corr{} $B=32$, Uniform\corr{} $B=32$, Uniform\corr{} $B=64$, and Uniform\corr{} $B=64$ in $\pi_{0.5}$ (whereas restricting the candidate set to a single suite-wide budget selects Uniform\corr{} $B=64$ in both models).

Held-out evaluation measures the selected configurations against Native on disjoint initial states 5--24 (800 conditions per model, 200 per suite) from the published initial states. On these held-out conditions, the selected configurations average 48 visual tokens per camera in $\pi_{0.5}$ and 40 in OpenVLA-OFT, trailing Native by 2.38 and 3.38 percentage points (2.375 and 3.375 unrounded).

\subsection{Independent LIBERO-Plus Confirmation}
\label{app:plus_confirmation}
\begin{table}[htbp]
\centering\small
\caption{Prespecified LIBERO-Plus comparisons in $\pi_{0.5}$ across 1,120 paired conditions at $B=16$ (task-cluster intervals over the 40 base tasks within four suites, with discordant success counts in the last column).}
\label{tab:plus_contrasts}
\begin{tabular}{@{}lrrrr@{}}
\toprule
Contrast & Confidence & $\Delta$ (pp) & Interval (pp) & Only first / second \\
\midrule
Tree\corr{} $-$ Uniform\corr{} & 95\% & $+5.63$ & $[2.41,8.93]$ & 163 / 100 \\
Tree\corr{} $-$ Native & 97.5\% & $-38.30$ & $[-44.91,-32.05]$ & 17 / 446 \\
Uniform\corr{} $-$ Native & 97.5\% & $-43.93$ & $[-51.34,-36.96]$ & 12 / 504 \\
\bottomrule
\end{tabular}

\end{table}
\begin{table}[htbp]
\centering\small
\caption{LIBERO-Plus success counts in $\pi_{0.5}$ at $B=16$ across all seven perturbation categories (160 paired conditions per category).}
\label{tab:plus_categories}
\begin{tabular}{@{}lrrrr@{}}
\toprule
Perturbation & Conditions & Native & Uniform\corr{} & Tree\corr{} \\
\midrule
Background Textures & 160 & 151/160 & 81/160 & 97/160 \\
Camera Viewpoints & 160 & 121/160 & 41/160 & 57/160 \\
Language Instructions & 160 & 139/160 & 72/160 & 69/160 \\
Light Conditions & 160 & 153/160 & 97/160 & 101/160 \\
Objects Layout & 160 & 134/160 & 82/160 & 99/160 \\
Robot Initial States & 160 & 117/160 & 45/160 & 52/160 \\
Sensor Noise & 160 & 144/160 & 49/160 & 55/160 \\
\bottomrule
\end{tabular}

\end{table}
\paragraph{Fixed Conditions and Pairing.}
From the seven perturbation categories of LIBERO-Plus~\cite{liberoplus} over the 40 LIBERO tasks, we fix four catalog variants per task and category prior to evaluation (excluding development variants and duplicate effective conditions), yielding 1,120 paired conditions. Across $B\in\{16,32,64\}$, all five configurations in Table~\mainref{tab:plus} (Native, Uniform, Uniform\corr{}, Tree, and Tree\corr{}, 5,600 episodes per budget) run on the same 1,120 conditions with environment seed 7, condition-indexed flow-noise schedules, ten flow steps, horizon ten, and five executed actions per request. Initial physical states and nonvisual inputs match identically across configurations, and rendered initial frames agree within at most one 8-bit level (offscreen renderer rounding).

\paragraph{Prespecified Comparisons and Budget Sweep.}
Resampling ten whole base tasks per suite across 10,000 bootstrap draws (seed 20261026) at $B=16$, Table~\ref{tab:plus_contrasts} reports the primary contrast (Tree\corr{} minus Uniform\corr{}, 95\% interval) and two secondary Native-reference contrasts (97.5\% intervals). Tree\corr{} reaches 47.32\% (530/1,120) versus 41.70\% (467/1,120) for Uniform\corr{} ($+5.63$ percentage points, 95\% interval $[2.41,8.93]$, 63 net successes on 163 vs.\ 100 discordant conditions, category breakdown in Table~\ref{tab:plus_categories}), while Native, Uniform, and Tree achieve 85.63\% (959/1,120), 3.39\% (38/1,120), and 6.88\% (77/1,120). At $B=32$, Native achieves 85.36\% (956/1,120), Uniform achieves 44.11\% (494/1,120) and 63.21\% (708/1,120) without and with the count bias, and Tree achieves 56.16\% (629/1,120) and 71.52\% (801/1,120). At $B=64$, Native achieves 85.54\% (958/1,120), Uniform achieves 75.45\% (845/1,120) and 79.73\% (893/1,120), and Tree achieves 78.04\% (874/1,120) and 80.54\% (902/1,120), completing Table~\mainref{tab:plus}.

\subsection{Paired LIBERO Episode}
\label{app:episode}
Figure~\ref{fig:episode} traces one paired $\pi_{0.5}$ rollout from start to finish: from the same initial state, adding the count bias places both objects in the basket, whereas uncorrected merging fails to release the first object and times out.
\begin{figure}[htbp]
\centering
\captionsetup{skip=4pt}
\includegraphics[width=0.95\columnwidth]{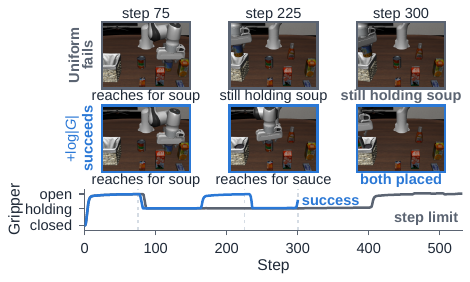}
\caption{\textbf{Same initial state, different readout.} $\pi_{0.5}$, Uniform at $B=64$, LIBERO-Long task ``put both the alphabet soup and the tomato sauce in the basket'' (bottom: gripper opening). With the count bias the soup is released at step 167, whereas without it the soup is held until step 407 and the episode times out. Selected in task and initial-state order as the first LIBERO-Long condition of the correction-path panel (Sec.~\ref{app:readout_closed_loop}) where uncorrected merging fails and dual-path correction succeeds (dual-path correction wins 3 of 4 discordant conditions in LIBERO-Long, and 9 of 13 across all four suites).}
\label{fig:episode}
\end{figure}

\subsection{Recorded Two-Camera Spatial Partitions}
\label{app:rgb_views}
\begin{figure}[htbp]
\centering
\captionsetup{skip=4pt}
\includegraphics[width=0.82\textwidth]{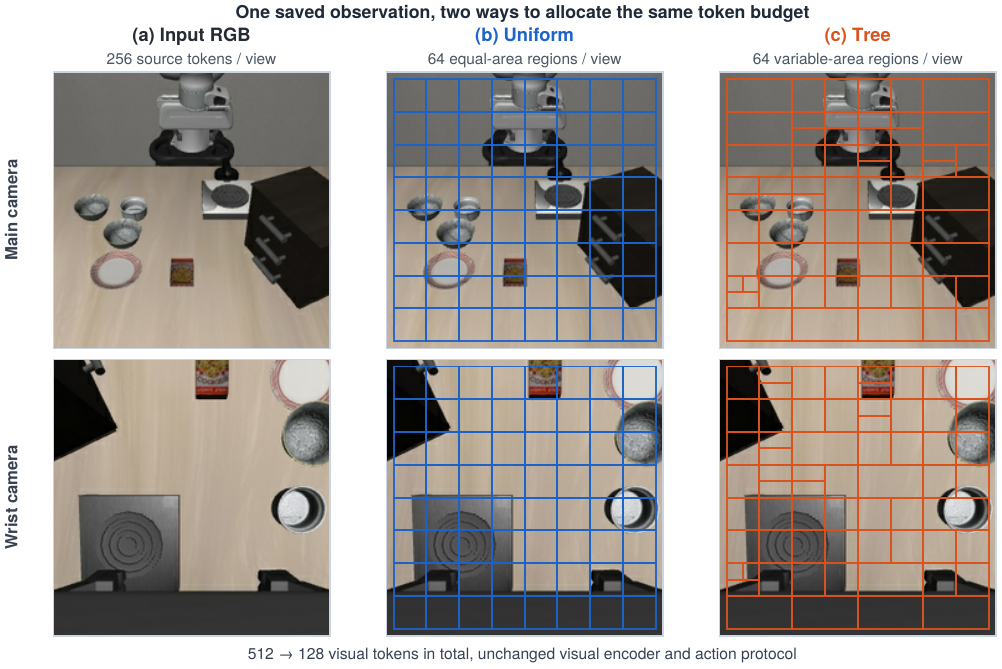}
\caption{\textbf{Recorded RGB inputs and spatial partitions for both cameras.}
LIBERO-Spatial task 0, initial state 0, first request: pick up the black bowl between the plate and the ramekin and place it on the plate.
Rows show the main and wrist cameras, column (a) shows the saved RGB input, and columns (b,c) overlay the recorded Uniform and Tree partitions under OpenVLA-OFT's 90\%-area center crop.}
\label{fig:rgb_views}
\end{figure}
Figure~\ref{fig:rgb_views} illustrates the recorded Uniform and Tree partitions at $B=64$ on the first archived RGB input (task 0, initial state 0 of LIBERO-Spatial) in OpenVLA-OFT. At the same budget of 64 tokens per view, Uniform assigns 4 patches to every region, whereas Tree adapts granularity between 1 and 8 patches around task-relevant objects and the gripper while leaving the underlying RGB input unchanged.

\section{Real-Robot Evaluation}
\label{supp:robot}

\subsection{Protocol and Outcomes}
\label{app:robot}
\paragraph{Setup.}
One $\pi_{0.5}$ checkpoint, LoRA-fine-tuned for one arm of a Galaxea R1 Lite robot and then frozen, serves both tasks. Native and the three compressed configurations share this checkpoint and one compiled runtime on an RTX~5090 policy server. Each query takes the front and wrist RGB images ($640\times480$), the seven-dimensional robot state (six arm joints and the gripper), and the language instruction, returning an action chunk of 50 seven-dimensional targets executed at 30\,Hz. All compressed configurations retain 64 tokens per camera, with the two $+\log|G|$ configurations applying the count bias along both readout paths of $\pi_{0.5}$.

\paragraph{Tasks.}
\emph{Cube into basket} (``put the rubik's cube into the brown basket'') executes 15 targets per query and terminates upon operator-recorded completion or 1,200 executed steps (40\,s at 30\,Hz), succeeding when the cube rests fully inside the basket with the gripper open. \emph{Block onto box} (``put the orange block on top of the green box'') executes 25 targets per query up to 2,000 executed steps (about 67\,s at 30\,Hz), succeeding when the released block rests stably on top of the box for at least 2\,s after the gripper clears.

\paragraph{Blocks.}
For each task, 50 paired blocks place the object at marked start position A or B (25 blocks each) and evaluate Native (256 tokens per camera) and the three $B=64$ configurations once per block under the shared checkpoint and success rule, with within-block order shuffled by a fixed seed on the block task.

\paragraph{Outcomes.}
Final outcomes are verified from saved front and wrist frames against control and event logs under the single task rule: on the cube task, step-limit episodes count as successes if the cube was released inside the basket with the gripper open before the limit, and on the block task, the final saved frames verify stable placement. Table~\ref{tab:robot_supp} reports both verified and operator-logged (Raw) counts, and every paired two-sided exact McNemar test (Holm-adjusted across the four comparisons per task) reaches the same 0.05-level significance decision under either count. On the block task, Uniform and Uniform\corr{} lift the block in only 1 of 100 combined trials, compared with 43/50 for Tree\corr{} and 37/50 for Native.

\paragraph{Readout.}
Sec.~\ref{app:r1lite} replays saved real-robot observations through the deployed checkpoint to measure action-path visual readout and joint-command error (Fig.~\mainref{fig:robot}c).

\paragraph{Latency.}
On the cube task, request latency is the median end-to-end server time (image decoding, preprocessing, token merging, and policy inference) across 16 saved observations replayed twice in interleaved order on the RTX~5090 server (with model forward alone taking 76.9\,ms for Native and 54.3\,ms for Tree\corr{}). On the block task, request latency is the mean server-side request time across all rollout requests of the 50 episodes (1,788 requests for Native and 1,836 for Tree\corr{}, with model forward alone taking 77.0 and 54.6\,ms).

\begin{table}[htbp]
\centering\footnotesize
\captionsetup{skip=4pt}
\caption{\textbf{Real-robot results.} Successes over 50 paired blocks per task (25 at each start position A and B), operator-logged counts (Raw), request latency (cube: replay median, block: rollout mean), and within-block discordant pairs with Holm-adjusted exact McNemar $p$-values.}
\label{tab:robot_supp}
\setlength{\tabcolsep}{3pt}
\adjustbox{max width=\linewidth}{
\begin{tabular}{lrrrrr}\toprule
Configuration & Success & A & B & Raw & Request\\\midrule
\multicolumn{6}{l}{\textit{Cube into basket}}\\
Native & 45/50 & 24/25 & 21/25 & 41/50 & 80.3\,ms\\
Uniform & 1/50 & 1/25 & 0/25 & 1/50 & 59.5\,ms\\
Uniform\corr{} & 30/50 & 11/25 & 19/25 & 28/50 & 58.5\,ms\\
Tree\corr{} & 45/50 & 23/25 & 22/25 & 44/50 & 58.5\,ms\\
\midrule\multicolumn{6}{l}{\textit{Block onto box}}\\
Native & 36/50 & 18/25 & 18/25 & 34/50 & 80.9\,ms\\
Uniform & 0/50 & 0/25 & 0/25 & 0/50 & 59.2\,ms\\
Uniform\corr{} & 0/50 & 0/25 & 0/25 & 0/50 & 58.8\,ms\\
Tree\corr{} & 40/50 & 17/25 & 23/25 & 41/50 & 59.1\,ms\\
\midrule
Paired comparison & \multicolumn{2}{r}{Cube} & \multicolumn{3}{r}{Block}\\
Uniform\corr{} vs.\ Uniform & \multicolumn{2}{r}{29\,:\,0, $<0.0001$} & \multicolumn{3}{r}{0\,:\,0, 1.0}\\
Tree\corr{} vs.\ Uniform\corr{} & \multicolumn{2}{r}{17\,:\,2, 0.002} & \multicolumn{3}{r}{40\,:\,0, $<0.0001$}\\
Tree\corr{} vs.\ Native & \multicolumn{2}{r}{4\,:\,4, 1.0} & \multicolumn{3}{r}{10\,:\,6, 0.909}\\
Uniform\corr{} vs.\ Native & \multicolumn{2}{r}{3\,:\,18, 0.003} & \multicolumn{3}{r}{0\,:\,36, $<0.0001$}\\
\bottomrule
\end{tabular}
}
\end{table}

\paragraph{Keyframes.}
For each task, Fig.~\mainref{fig:overview} displays the first paired block in run order matching the task's modal outcome pattern across its 50 blocks (Native, Uniform\corr{}, and Tree\corr{} succeeding while Uniform fails in 25/50 cube blocks, and Native and Tree\corr{} succeeding while both Uniform configurations fail in 30/50 block-task blocks, both starting at position A). The top row renders the wrist view nearest to 20\,s of the cube-task Tree\corr{} trial by region mean RGB color ($16\times16$ Native patches, $2\times2$ Uniform groups, or the recorded Tree\corr{} partition), dimmed in proportion to relative visual share. The middle row plots mean absolute and relative visual shares across the 200 replayed inputs per task (Sec.~\ref{app:r1lite}), and the bottom row shows identically cropped final front-camera frames.

\subsection{Readout Diagnostics on the Real-Robot Policy}
\label{app:r1lite}
\paragraph{Paired Inputs.}
We replay saved observations from both real-robot tasks (Sec.~\ref{app:robot}) through the deployed frozen $\pi_{0.5}$ checkpoint. Each of the 50 Native trials per task contributes the four saved requests nearest to 12.5\%, 37.5\%, 62.5\%, and 87.5\% of its request span, yielding 200 inputs per task (100 per start position), all evaluated with identical images, arm states, instructions, and initial flow-noise seeds across Native, Uniform, Uniform\corr{}, and Tree\corr{} at $B=64$.

\paragraph{Metrics.}
Across all 10 flow steps, 18 action-expert layers, and 8 heads, we average over the 50 action queries both the visual attention share and the 256-channel visual value contribution, comparing each compressed configuration against same-input Native via visual log-odds shift, absolute share error, visual-message RMSE, and six-joint arm command RMSE (over all 50 predicted targets, in radians) along each configuration's own flow trajectory. Observations are averaged within trials first, weighting trials equally, with 95\% intervals from 10,000 trial-level bootstrap draws stratified by start position.

\paragraph{Checks.}
Attaching the eager attention observer leaves predicted actions bitwise identical on fixed check inputs, and eager Native differs from the compiled deployment runtime by only 0.0026 and 0.0019\,rad in joint-command RMSE on the two tasks, an order of magnitude smaller than the compression errors in Table~\ref{tab:r1lite_mechanism}.

\begin{table}[!tb]
\centering\footnotesize
\captionsetup{skip=4pt}
\caption{\textbf{Readout diagnostics on saved real-robot inputs.} Trial-weighted means across 200 replayed inputs per task at $B=64$, reporting action-query visual attention share and differences from same-input Native (visual log-odds shift, share error (pp), visual-message RMSE, and joint-command RMSE).}
\label{tab:r1lite_mechanism}
\setlength{\tabcolsep}{3pt}
\adjustbox{max width=\linewidth}{
\begin{tabular}{lrrrrr}\toprule
Configuration & Visual share & Log-odds shift & Share error & Message & Joint (rad)\\\midrule
\multicolumn{6}{l}{\textit{Cube into basket}}\\
Native & 15.5\% & -- & -- & -- & --\\
Uniform & 6.0\% & $-1.41$ & 9.6 & 0.083 & 0.057\\
Uniform\corr{} & 13.6\% & $-0.29$ & 2.7 & 0.062 & 0.038\\
Tree\corr{} & 14.1\% & $-0.18$ & 2.3 & 0.058 & 0.034\\
\midrule\multicolumn{6}{l}{\textit{Block onto box}}\\
Native & 16.1\% & -- & -- & -- & --\\
Uniform & 6.1\% & $-1.45$ & 10.1 & 0.088 & 0.054\\
Uniform\corr{} & 13.9\% & $-0.32$ & 3.1 & 0.068 & 0.035\\
Tree\corr{} & 14.7\% & $-0.19$ & 2.3 & 0.060 & 0.023\\
\bottomrule
\end{tabular}
}
\end{table}

\paragraph{Results.}
With four patches per region, the count term predicts a $-\log 4=-1.39$ visual log-odds shift under uncorrected Uniform, closely matching the measured $-1.41$ (cube) and $-1.45$ (block) shifts in Table~\ref{tab:r1lite_mechanism}. Adding the count bias reduces the shift to $-0.29$ and $-0.32$, cutting share error by 71.7\% and 69.6\% (95\% intervals $[71.3,72.0]$ and $[69.2,69.9]$), message error by 24.6\% and 22.0\%, and joint-command error by 32.5\% and 34.6\% ($[27.4,37.0]$ and $[31.2,37.9]$). Replacing Uniform with Tree further narrows the shift to $-0.18$ and $-0.19$ and lowers message error by an additional 7.0\% and 12.1\% and joint-command error by 11.4\% and 34.1\% ($[4.6,17.6]$ and $[29.7,38.8]$), with the largest gain on the block task where count correction alone succeeds in 0/50 trials.
\endgroup

\end{document}